\documentclass[11pt]{article}

\usepackage{graphicx} % For images
\usepackage{amsmath, amssymb} % For math symbols
\usepackage{algorithm}
\usepackage{float}
\usepackage{algorithmic}
\usepackage[numbers]{natbib} % Citation style compatible with arXiv
\usepackage{authblk} % For author and affiliation formatting
\usepackage[margin=1in]{geometry} % Adjust margins for arXiv
\usepackage{hyperref}
\usepackage{url}
\usepackage{booktabs}
\usepackage{multirow}
\newtheorem{definition}{Definition}
\usepackage{mathtools}

\title{Federated Learning of Dynamic Bayesian Network via Continuous Optimization from Time Series Data}
\author[1]{Jianhong Chen}
\author[2]{Ying Ma}
\author[1]{Xubo Yue\thanks{Corresponding Author: \texttt{x.yue@northeastern.edu}}}

\affil[1]{Department of Mechanical \& Industrial Engineering, Northeastern University, Boston, MA, USA}
\affil[2]{Center for Computational Molecular Biology \& Department of Biostatistics, Brown University, Providence, RI, USA}
\begin{document}
\maketitle
\begin{abstract}
Traditionally, learning the structure of a Dynamic Bayesian Network has been centralized, requiring all data to be pooled in one location. However, in real-world scenarios, data are often distributed across multiple entities (e.g., companies, devices) that seek to collaboratively learn a Dynamic Bayesian Network while preserving data privacy and security. More importantly, due to the presence of diverse clients, the data may follow different distributions, resulting in data heterogeneity. This heterogeneity poses additional challenges for centralized approaches. In this study, we first introduce a federated learning approach for estimating the structure of a Dynamic Bayesian Network from homogeneous time series data that are horizontally distributed across different parties. We then extend this approach to heterogeneous time series data by incorporating a proximal operator as a regularization term in a personalized federated learning framework. To this end, we propose \texttt{FDBNL} and \texttt{PFDBNL}, which leverage continuous optimization, ensuring that only model parameters are exchanged during the optimization process. Experimental results on synthetic and real-world datasets demonstrate that our method outperforms state-of-the-art techniques, particularly in scenarios with many clients and limited individual sample sizes.

\end{abstract}

\section{Introduction}

Learning the structure of a \textbf{Dynamic Bayesian Network} (DBN) is a fundamental technique in causal learning and representation learning. DBNs are particularly valuable for uncovering temporal dependencies and conditional independence relationships in time-series data, making them a cornerstone of causal inference in dynamic systems. More importantly, DBNs have been widely applied in real-world domains, such as modeling gene expression \citep{lemoine2021gwena}, assessing network security \citep{Chock2017}, analyzing manufacturing process parameters \citep{SUN2020577}, and recognizing interaction activities \citep{zeng2016discovering}. 

Traditionally, DBN structure learning has been conducted in a centralized setting, where all data is aggregated in one location. However, with the rapid advancements in technology and the proliferation of Internet of Things (IoT) \citep{Madakam2015IoT} devices, data collection has become significantly more decentralized. In real-world scenarios, data is often owned and distributed across multiple entities, such as mobile devices, individuals, companies, and healthcare institutions. Many of these entities, referred to as clients, may lack a sufficient number of samples to independently construct a meaningful DBN. To gain comprehensive insights, they may seek to collaboratively learn the DBN structure while preserving data privacy and security. For instance, consider several hospitals aiming to collaboratively develop a DBN to uncover conditional independence relationships among medical conditions. Sharing patient records is strictly prohibited by privacy regulations \citep{hipaa_privacy_rule}, making centralized data aggregation infeasible. Furthermore, data heterogeneity introduces additional complexities: clients may exhibit homogeneous data (similar statistical distributions) or heterogeneous data, where distributions differ significantly or feature spaces (e.g., variables or attributes) vary entirely. These disparities complicate collaborative learning, as models must reconcile mismatched patterns or structural differences without direct data sharing.

\textit{A robust privacy-preserving learning strategy is therefore essential.} In recent years, federated learning (FL) \citep{McMahan2016CommunicationEfficientLO} has gained significant attention as a framework that enables multiple clients to collaboratively train machine learning models in a privacy-preserving manner. In FL, clients do not share raw data but instead exchange minimal necessary information, such as model parameters and gradient updates, relevant to the learning task. This approach has been successfully applied across various domains, including the Internet of Things \citep{Kontar2021Ioft}, condition monitoring \citep{arabi2024federated, zhang2024federated}, and recommender systems \citep{Chai2020SecureFederated}. Readers seeking further insights into FL can refer to comprehensive review papers \citep{Li2020FederatedLearning, Yang2019FederatedMachine}. Moreover, Personalized Federated Learning (PFL) extends FL to address the challenges posed by heterogeneous data distributions across clients. Unlike traditional FL, which trains a single global model, PFL allows for the development of personalized models tailored to each client’s unique data distribution while still benefiting from collaborative learning. To achieve this balance between personalization and collaboration, various PFL approaches have been proposed, including proximal regularization \citep{Sahu2018FederatedOI} to balance local and global models, clustered federated learning \citep{GhoshCFL2022, Sattler2020CFL, Muhammad2020FedFast} to group clients with similar data distributions, and meta-learning-based methods \citep{Jiang2019metaFL, Chen2018FederatedMW} to facilitate fast local adaptation. These techniques enhance performance in real-world federated environments while preserving data privacy and minimizing communication costs.

Traditionally, score-based DBN learning relies on discrete optimization, making it incompatible with common continuous optimization techniques. However, continuous optimization methods, particularly those used in federated learning, solve problems efficiently using convex optimization techniques. Recent works by \citet{Zheng2018} and \citet{pamfil2020dynotears} have addressed this limitation by introducing continuous optimization strategies for DBN learning, leveraging algebraic characterizations of acyclicity. Despite this advancement, directly applying popular federated optimization techniques - such as FedAvg \citep{McMahan2016CommunicationEfficientLO} and Ditto \citep{tian2021ditto} - to these continuous formulations remains challenging due to the inherent acyclicity constraint in DBN structure learning.

\paragraph{Contributions.} In this work, we introduce a federated learning approach and a personalized federated learning approach for estimating the structure of a Dynamic Bayesian Network from homogeneous and heterogeneous observational time-series data that is horizontally partitioned across different clients. Our contributions can be summarized as follows:

\begin{itemize}
    \item \textbf{Novel Federated Approach for DBN Learning:}  
    \begin{itemize}
        \item \textbf{Federated DBN Learning (\texttt{FDBNL})} for homogeneous time series, formulated as a continuous optimization problem and solved using the \textbf{Alternating Direction Method of Multipliers (ADMM)}.  
        \item \textbf{Personalized Federated DBN Learning (\texttt{PFDBNL})} for heterogeneous time series, incorporating a \textbf{proximal operator} to account for client-specific variations.  
    \end{itemize}
    
    \item Both \texttt{FDBNL} and \texttt{PFDBNL} enable \textbf{causal structure learning} directly from data without imposing restrictive assumptions on the underlying graph topology.  

    \item Through extensive simulations and experiments on two high-dimensional, real-world datasets, we demonstrate that our methods are \textbf{scalable} and effectively recover the true underlying DBN structures, even in challenging heterogeneous settings.  

    % \item We outline several promising \textbf{research directions} for future work, as discussed in \S~\ref{dis}.  

    \item To the best of our knowledge, this is the \textbf{first study} to explore DBN structure learning within the \textbf{federated learning} framework. Additionally, we pioneer the study of \textbf{temporal dynamic structure learning from heterogeneous data} in a federated context.  
\end{itemize}

Our full code is available on GitHub: \url{https://github.com/PeChen123/FDBN_Learning}.

\paragraph{Literature Review}
Dynamic Bayesian Networks are probabilistic models that extend static Bayesian Networks to represent multi-stage processes by modeling sequences of variables across different time steps. Learning a DBN involves two key tasks: \textit{structure learning} (identifying the network topology) and \textit{parameter learning} (estimating the conditional probabilities). \citet{Firedman2013} laid the foundation for this field by developing methods to learn the structure of dynamic probabilistic networks, effectively extending Bayesian Networks to capture temporal dynamics in sequential data. Building on this work, \citet{Murphy2002} adapted static Bayesian Network learning techniques for temporal domains, employing methods such as the Expectation-Maximization algorithm for parameter estimation. Advancements in structure learning from time-series data were further propelled by \citet{pamfil2020dynotears}, who developed an extension of Bayesian Network inference using continuous optimization methods. Their approach enabled efficient and scalable learning of DBN structures by formulating the problem as a continuous optimization task with acyclicity constraints tailored for temporal graphs. To address non-linear dependencies in time-series data, \citet{Tank2022} introduced neural Granger causality methods, which leverage neural networks within the DBN framework to capture complex non-linear interactions. In the field of bioinformatics, \citet{Yu2004} enhanced Bayesian Network inference techniques to generate causal networks from observational biological data. Their work demonstrated the practical applicability of DBNs in investigating intricate biological relationships and regulatory mechanisms. However, traditional DBN learning methods often require centralized data aggregation, which poses significant privacy risks and scalability challenges, particularly in sensitive domains such as healthcare and genomics. These limitations have spurred interest in privacy-preserving and distributed learning frameworks. Federated Learning enables DBNs to be trained collaboratively across distributed datasets, facilitating secure and scalable model inference while preserving data privacy. 

\textbf{To the best of our knowledge, there is very limited work on federated or privacy-preserving approaches for learning Dynamic Bayesian Networks.} However, there are numerous studies on distributed learning for Bayesian Networks. \citet{Guo2007} adopted a two-step procedure that first estimates the BN structures independently using each client’s local dataset and then applies a further conditional independence test. \citet{Na2010DistributedBN} proposed a voting mechanism to select edges that were identified by more than half of the clients. These approaches rely solely on the final graphs independently derived from each local dataset, which may lead to suboptimal performance due to the limited exchange of information. More importantly, \citet{ng2022federated} proposed a distributed Bayesian Network learning method based on continuous optimization using the Alternating Direction Method of Multipliers (ADMM). Notably, no prior work has addressed distributed learning of DBNs.

\section{Problem Statement}

We consider a scenario with \( K \) clients, indexed by \( k \in \{1, 2, \ldots, K\} \). Each client \( k \) possesses a local dataset containing \( M \) realizations of a stationary time series. Specifically, for the \( k \)-th client, the time series dataset is denoted as \(\mathcal{D}^k = \{ x_{m,t}^{k} \}_{t=0}^{T}\), where \( x_{m,t}^{k} \in \mathbb{R}^d \), \( m \in \{1, 2, \ldots, M\} \) indexes realizations, and \( d \) is the number of variables. Due to privacy constraints, these time series cannot be directly shared between clients. For simplicity, we omit the realization subscript \( m \) in the main text (see supplementary material \S\ref{m_not} for full notation) and consider a generic realization \(\{ x_{t}^k \}_{t=0}^{T}\).

For each client \( k \), we model the data \(\{x_t^k\}\) (for a single realization) using a Structural Equation Model (SEM) with a Structural Vector Autoregressive (SVAR) structure of order \( p \):

\[
\label{data}
(x_t^k)^\top = (x_t^k)^\top W_k + (x_{t-1}^k)^\top A_{k_1} + \dots + (x_{t-p}^k)^\top A_{k_p} + (u_t^k)^\top,
\]

where:
\begin{enumerate}
    \item \( t \in \{ p, p+1, \ldots, T \} \), and \( p \) is the autoregressive order.
    \item \( u_t^k \sim \mathcal{N}(0, I) \) is a noise vector, independent across time and variables.
    \item \( W_k \) is a weighted adjacency matrix (structured as a directed acyclic graph (DAG)) capturing intra-slice dependencies for the \( k \)-th client.
    \item \( A_{k_i} \) (\( i \in \{1, \dots, p\} \)) are coefficient matrices capturing inter-slice dependencies for the \( k \)-th client.
\end{enumerate}

The matrices \( W_k \) and \( A_{k_i} \) may be homogeneous (shared across all clients) or heterogeneous (client-specific). We assume the order \( p \) is same for all clients. In the homogeneous case, we write \( W_k = W \) and \( A_{k_i} = A_i \) since \(W_k\) and \( A_{k_i} \) are the same for all clients. In the heterogeneous case, these matrices vary by client. The sample size for client \( k \) is \( n_k = T + 1 - p \).

Given the federated dataset \( X = \bigcup_{k=1}^K \mathcal{D}^k \), the goal is to recover the true parameters \( W, A_i \) (homogeneous) or \( W_k, A_{k_i} \) (heterogeneous) in a privacy-preserving manner. Clients collaborate to learn the DBN structure, combining intra-slice (\( W_k \)) and inter-slice (\( A_{k_i} \)) dependencies, while sharing only minimal information (e.g., model parameters or graphs). The total sample size is \( n = \sum_{k=1}^K n_k \). We also consider partial client participation, where only a subset of clients \( j \leq K \) contributes in each communication round. This is critical for scalability, as inactive clients can rejoin later without disrupting the process. Notably, partial participation is primarily relevant in the heterogeneous case, where client-specific parameters require updates from participating clients. In the homogeneous case, shared parameters remain unaffected by non-participating clients.

\section{Proposed Methodologies}
\begin{figure}[H]
\centering
\includegraphics[width=\textwidth]{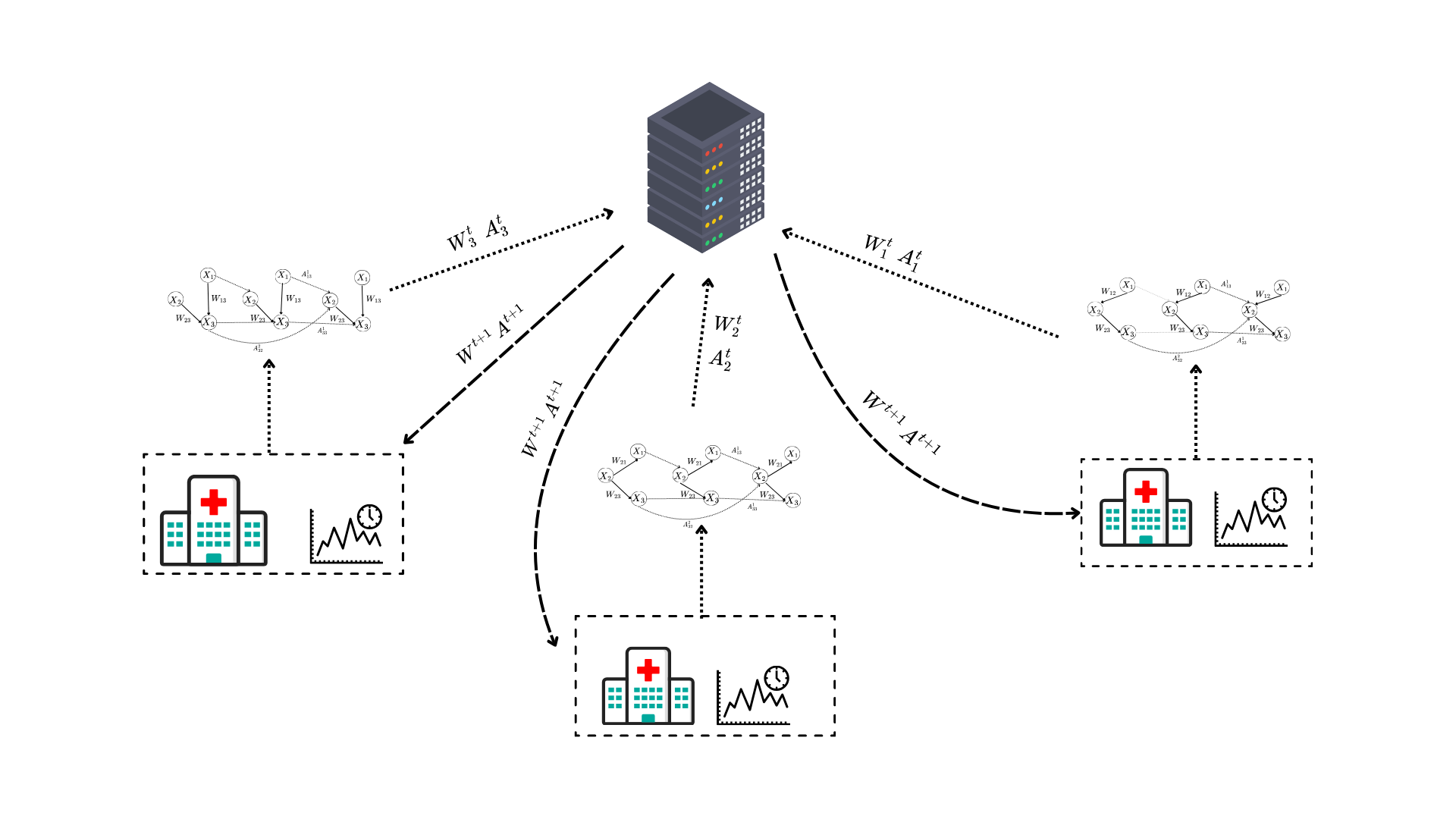}
\caption{Overview of Personalized Federated DBN learning with \( d = 3 \) nodes, autoregression order \( p = 2 \), and \( K = 3 \) clients. The only difference between it and Federated DBN learning is that Federated DBN learning requires each \( W_k \) and \( A_{k_i} \) to be identical across clients.} 
\label{over}
\end{figure}

In this section, we introduce the framework for Federated Dynamic Bayesian Network Learning and Personalized Federated Dynamic Bayesian Network Learning. The general idea is summarized in Figure~\ref{over}.

\subsection{Federated Dynamic Bayesian Network Learning}
\label{method}
In this section, we introduce the Federated DBN learning (\texttt{FDBNL}) for the homogeneous (i.i.d) time series. To implement federated learning to DBN, our method builds on the method introduced by \citet{pamfil2020dynotears}, which frames the structural learning of linear DBN as a continuous constrained optimization problem. By incorporating \( \ell_1 \)-norm penalties to encourage sparsity, the optimization problem is formulated as follows:

\begin{equation}
\begin{split}
   \min_{W,A} &\ \ell(W,A) + \lambda_W \| W \|_1 + \lambda_A \| A \|_1 \quad \text{subject to} \quad W \text{ is acyclic}, \\
   &\ell(W,A) = \frac{1}{2n} \| X_t - X_t W - X_{(t-p):(t-1)} A \|_F^2 ,
\end{split}
\label{V1}
\end{equation}
where \( X_{(t-p):(t-1)} = \left[ X_{t-1} \ \cdots \ X_{t-p} \right] \), with each \( X_{t-i} \) (for \( i \in \{1, \ldots, p\} \)) being a time-lagged version of $X_t$ and \( n \times d \) matrix where rows correspond to realizations. The matrix \( A \) is defined as:
\[
A = \begin{bmatrix} A_1^\top \\ \vdots \\ A_p^\top \end{bmatrix},
\]
where each \( A_i \in \mathbb{R}^{d \times d} \) captures lag-\(i\) dependencies. Thus, 
we observe that \( X_{(t-p):(t-1)} \) is an \( n \times (p \cdot d) \) matrix, while \( A \) is a \( (p \cdot d) \times d \) matrix. For the acyclicity constraint, following \citet{Zheng2018},  

\[
h(W) = \mathrm{tr}\left( e^{W \circ W} \right) - d
\]  
is equal to zero if and only if \( W \) is acyclic. Here, \( \circ \) represents the Hadamard product (element-wise multiplication) of two matrices. By replacing the acyclicity constraint with the equality constraint \( h(W) = 0 \), the problem can be reformulated as an equality-constrained optimization task.

This setting does not work for federated learning since all data is encapsulated. Thus, for the federated setting, we leverage the Alternating Direction Method of Multipliers \citep{Boyd2011ADMM}, an optimization algorithm designed to solve convex problems by decomposing them into smaller, more manageable subproblems. It is particularly effective for large-scale optimization tasks with complex constraints. By following ADMM framework, we decompose the constrained problem \eqref{V1} into multiple subproblems and utilize an iterative message-passing approach to converge to the final solution. ADMM proves particularly effective when subproblems have closed-form solutions, which we derive for the first subproblem. To cast problem \eqref{V1} in an ADMM framework, we reformulate it using local variables \( B_1, \dots, B_K \in \mathbb{R}^{d \times d} \) for the intra-slice matrices and \( D_1, \dots, D_K \in \mathbb{R}^{pd \times d} \) for the inter-slice matrices, along with shared global variables \( W \in \mathbb{R}^{d \times d} \) and \( A \in \mathbb{R}^{pd \times d} \), representing an equivalent formulation:

\begin{align*}
&\min_{B_k,D_k,W,A} \sum_{k=1}^K \ell_k(B_k,D_k) + \lambda_W \| W \|_1 + \lambda_A \| A \|_1\\
& \quad \text{subject to } h(W) = 0, \\
& \quad B_k = W, \quad k = 1,2,\ldots,K,\\
& \quad D_k = A, \quad k = 1,2,\ldots,K.
\end{align*}

The local variables \( B_1, \dots, B_K \) and \( D_1, \dots, D_K \) represent the model parameters specific to each client. The data for the $k$-th client consists $X_t^k$ and $X^k_{(t-p):(t-1)}$. Notably, this problem resembles the global variable consensus ADMM framework. The constraints \( B_k = W \) and \( D_k = A\) are imposed to enforce consistency, ensuring that the local model parameters across clients remain identical. Since ADMM combines elements of dual decomposition and the augmented Lagrangian method, making it efficient for handling separable objective functions and facilitating parallel computation, we employ the augmented Lagrangian method to transform the constrained problem into a series of unconstrained subproblems. The augmented Lagrangian is given by:

\begin{align*}
\mathcal{L}\left( \{ B_k, D_k \}_{k =1}^K, W, A, \alpha, \{ \beta_k, \gamma_k \}_{k =1}^K; \rho_1, \rho_2 \right) &=
\sum_{k=1}^K \Bigg[ \ell_k(B_k, D_k) + \text{tr}\left( \beta_k^\top (B_k - W) \right) + \frac{\rho_2}{2} \| B_k - W \|_F^2 \\
& \quad + \text{tr}\left( \gamma_k^\top (D_k - A) \right) + \frac{\rho_2}{2} \| D_k - A \|_F^2 \Bigg] \\
& + \lambda_W \| W \|_1 + \lambda_A \| A \|_1 + \alpha h(W) + \frac{\rho_1}{2} h(W)^2,
\end{align*}
where \( \{ \beta_k \}_{k =1}^K \in \mathbb{R}^{d \times d} \), \( \{ \gamma_k \}_{k =1}^K \in \mathbb{R}^{pd \times d} \) and \( \alpha \in \mathbb{R} \) are estimations of the Lagrange multipliers; \( \rho_1 \) and \( \rho_2 \) are the penalty coefficients and $\|\cdot\|_F$ is Frobenius norm. Then, we have the iterative update rules of ADMM as follows:

\textbf{Local Updates for \( B_k \) and \( D_k \)}

\begin{equation}
\label{Local}
\begin{split}
    (B_k^{(t+1)}, D_k^{(t+1)}) = & \arg \min_{B_k, D_k} \bigg[ \ell_k(B_k, D_k) + \text{tr}\left( \beta_k^{(t)\top} (B_k - W^{(t)}) \right) \\
    & + \frac{\rho_2^{(t)}}{2} \| B_k - W^{(t)} \|_F^2  +  \text{tr}\left( \gamma_k^{(t)\top} (D_k - A^{(t)}) \right) + \frac{\rho_2^{(t)}}{2} \| D_k - A^{(t)} \|_F^2 \bigg].
\end{split}
\end{equation}

\textbf{Global Updates for \( W \) and \( A \)}

\begin{equation}
\label{Global}
\begin{split}
    (W^{(t+1)}, A^{(t+1)}) = & \arg \min_{W, A} \Bigg[ \alpha^{(t)} h(W) + \frac{\rho_1^{(t)}}{2} h(W)^2 + \lambda_W \| W \|_1 + \lambda_A \| A \|_1 \\
& + \sum_{k=1}^K \left( \text{tr}\left( \beta_k^{(t)\top} (B_k^{(t+1)} - W) \right) + \frac{\rho_2^{(t)}}{2} \| B_k^{(t+1)} - W \|_F^2 \right) \\
& + \sum_{k=1}^K \left( \text{tr}\left( \gamma_k^{(t)\top} (D_k^{(t+1)} - A) \right) + \frac{\rho_2^{(t)}}{2} \| D_k^{(t+1)} - A \|_F^2 \right) \Bigg].
\end{split}
\end{equation}

\textbf{Update Dual Variables}

\begin{equation}
\label{para}
\begin{split}
\beta_k^{(t+1)} & = \beta_k^{(t)} + \rho_2^{(t)} (B_k^{(t+1)} - W^{(t+1)}), \\
\gamma_k^{(t+1)} & = \gamma_k^{(t)} + \rho_2^{(t)} (D_k^{(t+1)} - A^{(t+1)}), \\
\alpha^{(t+1)} & = \alpha^{(t)} + \rho_1^{(t)} h(W^{(t+1)}), \\
\rho_1^{(t+1)} & = \phi_1 \rho_1^{(t)}, \\
\rho_2^{(t+1)} & = \phi_2 \rho_2^{(t)},
\end{split}
\end{equation}
where \( \phi_1 , \phi_2 \in \mathbb{R} \) are hyperparameters that control how fast the coefficients \( \rho_1, \rho_2 \) are increased. As previously mentioned, ADMM is particularly efficient when the optimization subproblems have closed-form solutions. The subproblem in equation \eqref{Local} is a well-known proximal minimization problem, extensively explored in the field of numerical optimization \citep{Combettes2011,Parikh2014}. For readability, we define the following matrices:

\[
\begin{aligned}
S &= \frac{1}{n_k} X_t^{k\top} X_t^k, \\
M &= \frac{1}{n_k} X_t^{k\top} X_{(t-p):(t-1)}^k, \\
N &= \frac{1}{n_k} X_{(t-p):(t-1)}^{k\top} X_{(t-p):(t-1)}^k, \\
P &= S + \rho_2^{(t)} I, \\
Q &= N + \rho_2^{(t)} I,
\end{aligned}
\]

and vectors:

\[
\begin{aligned}
b_1 &= S - \beta_k^{(t)} + \rho_2^{(t)} W^{(t)}, \\
b_2 &= M^\top - \gamma_k^{(t)} + \rho_2^{(t)} A^{(t)}.
\end{aligned}
\]

By computing the gradient, we can derive the following closed-form solutions:

\begin{itemize}
    \item \( B_k^{(t+1)} \):
   \[
   \boxed{ B_k^{(t+1)} = \left( P - M Q^{-1} M^\top \right)^{-1} \left( b_1 - M Q^{-1} b_2 \right). }
   \]
    \item \( D_k^{(t+1)} \):
   \[
   \boxed{ D_k^{(t+1)} = \left( Q - M^\top P^{-1} M \right)^{-1} \left( b_2 - M^\top P^{-1} b_1 \right). }
   \]
\end{itemize}

For a comprehensive procedure, we have summarized it in the supplementary material~\S\ref{closed form}. The presence of the acyclicity term \( h(W) \) prevents us from obtaining a closed-form solution for the problem in equation \eqref{Global}. Instead, the optimization problem can be addressed using first-order methods, such as gradient descent, or second-order methods, like L-BFGS \citep{Byrd2003}. In this work, we write $W = W_{+} - W_{-}$ (and analogously for $A$) to handle the $\ell_1$ penalty term and employ the L-BFGS method to solve the optimization problem. We have summarized our procedure in Algorithm~\ref{alg:distributed-dbnsl}.

\begin{algorithm}[H]
\caption{Federated Dynamic Bayesian Network Learning}
\label{alg:distributed-dbnsl}
\begin{algorithmic}[1]
\REQUIRE Initial parameters \( \rho_1, \rho_2, \alpha^{(1)}, \beta_1^{(1)}, \ldots, \beta_K^{(1)}, \gamma_1^{(1)}, \ldots, \gamma_K^{(1)} \); multiplicative factors \( \phi_1 , \phi_2 > 1 \); initial points \( W^{(1)} \) and \( A^{(1)} \)
\FOR{$t = 1, 2, \ldots$}
    \STATE Each client solves problem \eqref{Local} in parallel
    \STATE Central server collects \( B_1^{(t+1)}, \ldots, B_K^{(t+1)}, D_1^{(t+1)}, \ldots, D_K^{(t+1)} \) from all clients
    \STATE Central server solves problem \eqref{Global}
    \STATE Central server sends \( W^{(t+1)} \) and \( A^{(t+1)} \) to all clients
    \STATE Central server updates ADMM parameters \( \alpha^{(t+1)}, \rho_1^{(t+1)}, \rho_2^{(t+1)} \) according to Eq.\eqref{para}
    \STATE Each client updates \( \beta_k^{(t+1)}, \gamma_k^{(t+1)} \) according to Eq.\eqref{para}
\ENDFOR
\end{algorithmic}
\end{algorithm}

\subsection{Personalized Federated Dynamic Bayesian Network Learning}
In this section, we introduce Personalized Federated DBN Learning (\texttt{PFDBNL}) for heterogeneous time series data. This method is based on the proximal operator, which is used to reformulate Problem.~\eqref{V1}, as introduced by \cite{Neal2014Prox}. The formal definition of the proximal operator is:

\begin{definition}[Proximal Operator]
Let $\mathcal{X}$ denote a vector space with norm $\|.\|_{\mathcal{X}}$, The proximal operator is defined as
\[
prox_{\mu f}(X) = \arg\min_{Z}{\mu||Z - X||_\mathcal{X}^2 + f(Z)}
\]
\end{definition}

To formulate the objective function for \texttt{PFDBNL}, we let  
\begin{equation}
    f_k(W_k,A_k) = \ell_k(W_k, A_k) + \lambda_W \| W_k \|_1 + \lambda_A \| A_k \|_1
\end{equation}
for the \( k \)-th client. Note that we sometimes write $A_k = \left[ A^\top_{k_1} \ \cdots \ A^\top_{k_p}  \right]^\top$ for \texttt{PFDBNL}-related representations.
Then, we adopt the method described in \citep{Sahu2018FederatedOI}, where \( f_k \) is replaced with its proximal operator. The detailed structure of the resulting optimization problem is outlined as follows:

\begin{equation}
    \begin{split}
     F_k(W,A) \coloneqq prox_{\mu f}(W, A) &= \min_{W_k, A_{k}} \mu||W_k - W||^2_{F} + \mu||A_k - A||^2_{F} + f_k(W_k, A_k) \\
    &= \min_{W_k, A_k} \ell(W_k,A_k) + \mu||W_k - W||^2_{F} + \mu||A_k - A||^2_{F} + \lambda_W \| W_k \|_1 + \lambda_A \| A_k \|_1.    
    \end{split}
    \label{PFDBN}
\end{equation}

The hyperparameter \( \mu \) regulates the balance between the global model \( (W, A) \) and the personalized model \( (W_k, A_k) \). A larger \( \mu \) benefits clients with less reliable data by leveraging aggregated information from other clients, while a smaller \( \mu \) prioritizes personalization for clients that possess a significant amount of high-quality data. Therefore, we define the following optimization problem:

\begin{equation}
\label{bi-levelPFL}
\begin{split}
   \min_{W,A} &\ \sum_{k=1}^K F_k(W,A) \quad \text{subject to} \quad W \text{ is acyclic}, \\
   &F_k(W,A)= \min_{W_k, A_k} \ell(W_k,A_k) + \mu||W_k - W||^2_{F} + \mu||A_k - A||^2_{F} + \lambda_W \| W_k \|_1 + \lambda_A \| A_k \|_1. 
\end{split}
\end{equation}

Problem~\eqref{bi-levelPFL} forms a bi-level optimization problem, which is traditionally solved by iteratively applying first-order gradient methods. The lower-level problem is solved first to approximate \( W_i \), and this solution is then used in the upper-level problem to optimize \( W \), repeating the process until convergence. However, while this approach is straightforward, it often suffers from low solution accuracy and challenges in fine-tuning parameters such as the learning rate. To overcome these limitations, we propose a relaxed formulation of Problem~\eqref{bi-levelPFL} to improve both efficiency and accuracy.

\begin{equation}
\label{Jointmini}
\begin{split}
&\min_{\{W_k, A_k\}_{k =1}^K, W,A,} \sum_{k=1}^K \ell_k(W_k,A_k) + \mu (||W_k - W||^2_{F} + ||A_k - A||^2_{F}) +\lambda_W \| W_k \|_1 + \lambda_A \| A_k \|_1\\
& \quad \text{subject to } h(W_k) = 0. \\
\end{split}
\end{equation}

Our goal is to learn an optimal personalized model \( (W_k, A_k) \) for each client $k$ and an optimal global model \( (W, A) \) that jointly minimize Problem~\eqref{Jointmini}, ensuring a balance between personalization and global knowledge aggregation. Since ADMM is a primal-dual method known for its improved iteration stability and faster convergence compared to gradient-based methods, we propose using ADMM to solve Problem~\eqref{Jointmini}. To facilitate this, we introduce auxiliary variables \( \{ \tilde{W}_k, \tilde{A}_k \}_{k=1}^{K} \), allowing us to reformulate Problem~\eqref{Jointmini} into a separable structure by partitioning the variables into multiple blocks. Specifically, this transformation enables a more efficient optimization framework, as detailed below:
\begin{equation}
\label{PDBNL_final}
    \begin{split}
    &\min_{\{W_k, A_k,\tilde{W}_k,\tilde{A}_k\}_{k =1}^K, W,A,} \sum_{k=1}^K \ell_k(W_k,A_k) + \mu (||W_k - \tilde{W}_k||^2_{F} + ||A_k - \tilde{A}_k||^2_{F}) +\lambda_W \| W_k \|_1 + \lambda_A \| A_k \|_1\\
    & \quad \text{subject to } h(W_k) = 0, \\
    & \quad \tilde{W}_k = W \\
    & \quad \tilde{A}_k = A, \quad k = 1,2,\ldots,K.      
    \end{split}
\end{equation}

Here, \( \{\tilde{W}_k,\tilde{A}_k \}_{k=1}^{K} \) represents the local model of client \( k \). Problem~\eqref{PDBNL_final} remains equivalent to Problem~\eqref{Jointmini} in the sense that they share the same optimal solutions. Since Problem~\eqref{PDBNL_final} involves linear constraints and multiple block variables, we can solve it by employing the exact penalty method within the ADMM framework, which is well-suited for handling such optimization structures. Furthermore, we assume \( h(W_k) = 0 \) for all clients, as we need to enforce the DAG constraint across all clients. Similarly, we apply the augmented Lagrangian method to reformulate the problem into a series of unconstrained subproblems. To implement ADMM for this problem, we define the corresponding augmented Lagrangian function as follows:

\[
\begin{aligned}
&\mathcal{L}\Bigl(\{W_k, A_k, \tilde{W}_k, \tilde{A}_k\}_{k=1}^K,\, W,\, A,\, \{\beta_k, \gamma_k\}_{k=1}^K;\, \alpha,\, \rho_1,\, \rho_2,\, \mu\Bigr) = \\
&\quad
\sum_{k=1}^K 
\Bigl[
  \ell_k\bigl(W_k, A_k\bigr)
  \;+\;
  \mu\,\|W_k - \tilde{W}_k\|_F^2
  \;+\;
  \mu\,\|A_k - \tilde{A}_k\|_F^2
  \;+\;
  \lambda_W\,\|W_k\|_1
  \;+\;
  \lambda_A\,\|A_k\|_1 \\
&\qquad\quad
  +\; \mathrm{tr}\Bigl(\beta_k^\top\!(\tilde{W}_k - W)\Bigr)
  \;+\;
  \frac{\rho_2}{2}\,\|\tilde{W}_k - W\|_F^2
  \;+\;
  \mathrm{tr}\Bigl(\gamma_k^\top\!(\tilde{A}_k - A)\Bigr)
  \;+\;
  \frac{\rho_2}{2}\,\|\tilde{A}_k - A\|_F^2
\Bigr] \\
&\quad 
+ \; \alpha \, h(W_k) 
\;+\; \frac{\rho_1}{2}\, \bigl[h(W_k)\bigr]^2,
\end{aligned}
\]
where \( \{ \beta_k \}_{k =1}^K \in \mathbb{R}^{d \times d} \), \( \{ \gamma_k \}_{k =1}^K \in \mathbb{R}^{pd \times d} \) and \( \alpha \in \mathbb{R} \) are estimates of the Lagrange multipliers; \( \rho_1 \) and \( \rho_2 \) are penalty coefficients, and \( \|\cdot\|_F \) denotes the Frobenius norm. Note that we set a single pair of \( \lambda_W, \lambda_A \) for all clients, as a generally effective pair can be found for datasets of the same dimension using \texttt{FDBNL} and prior studies \citep{ng2022federated, pamfil2020dynotears, Zheng2018}. Due to the personalization problem, each \( W_k, A_k \) may have different dimensions or structural properties (e.g., connectivity). In this study, we consider only cases where 
all clients share identical dimensions and connectivity. However, extending the framework to handle different dimensions or varying connectivity structures is an interesting direction for future research, as discussed in \S~\ref{dis}. We now define the iterative update rules of ADMM. Given initial values for \( \{W_k, A_k, \tilde{W}_k, \tilde{A}_k\}_{k = 1}^K \), \( W \), \( A \), and the dual variables, the updates are as follows:

\textbf{For $W_k, A_k$}
\begin{equation}
\label{p_local1}
\begin{split}
(W_k^{(t+1)}, A_k^{(t+1)})
\;=\;
\arg\min_{W_k,A_k}
\Bigl[
 \ell_k(W_k,A_k)
 &+ \mu \|W_k - \tilde{W}_k^{(t)}\|^2
 + \mu \|A_k - \tilde{A}_k^{(t)}\|^2\\
 &  +\alpha^{(t)}\,h(W_k) + \tfrac{\rho_1^{(t)}}{2}\,[h(W_k)]^2  +\lambda_W \|W_k\|_1 + \lambda_A \|A_k\|_1
\Bigr].
\end{split}
\end{equation}

\textbf{For $\tilde{W}_k, \tilde{A}_k$}:

\begin{equation}
\label{p_local2}
\begin{split}
(\tilde{W}_k^{(t+1)}, \tilde{A}_k^{(t+1)})
\;=\;
\arg\min_{\tilde{W}_k,\tilde{A}_k}
\Bigl[
 & \mu\|W_k^{(t+1)} - \tilde{W}_k\|^2
 + \mu\|A_k^{(t+1)} - \tilde{A}_k\|^2
  + \mathrm{tr}\bigl(\beta_k^{(t)\top}(\tilde{W}_k - W^{(t)})\bigr)\\
& + \tfrac{\rho_2^{(t)}}{2}\|\tilde{W}_k - W^{(t)}\|^2 
 + \mathrm{tr}\bigl(\gamma_k^{(t)\top}(\tilde{A}_k - A^{(t)})\bigr)
 + \tfrac{\rho_2^{(t)}}{2}\|\tilde{A}_k - A^{(t)}\|^2
\Bigr].
\end{split}
\end{equation}

\textbf{For $W,A$}:

\begin{equation}
\label{p_cen}
    \begin{split}
    (W^{(t+1)}, A^{(t+1)})
   \;=\;
   &\arg\min_{W,A}
   \Bigl[
      \sum_{k=1}^K
       \Bigl(
         \mathrm{tr}\bigl(\beta_k^{(t)\top}(\tilde{W}_k^{(t+1)} - W)\bigr)
         + \tfrac{\rho_2^{(t)}}{2}\|\tilde{W}_k^{(t+1)} - W\|^2
       \Bigr)\\
     &+ \sum_{k=1}^K
       \Bigl(
         \mathrm{tr}\bigl(\gamma_k^{(t)\top}(\tilde{A}_k^{(t+1)} - A)\bigr)
         + \tfrac{\rho_2^{(t)}}{2}\|\tilde{A}_k^{(t+1)} - A\|^2
       \Bigr)
   \Bigr].
   \end{split}
\end{equation}

\textbf{For Dual variables}:
\begin{equation}
\label{para1}
\begin{split}
\beta_k^{(t+1)} & = \beta_k^{(t)} + \rho_2^{(t)} (\tilde{W}_k^{(t+1)} - W^{(t+1)}), \\
\gamma_k^{(t+1)} & = \gamma_k^{(t)} + \rho_2^{(t)} (\tilde{A}_k^{(t+1)} - A^{(t+1)}), \\
\alpha^{(t+1)} & = \alpha^{(t)} + \rho_1^{(t)}  (\frac{1}{K}\sum_{k =1 }^{K} h(W_k^{(t+1)})),, \\
\rho_1^{(t+1)} & = \phi_1 \rho_1^{(t)}, \\
\rho_2^{(t+1)} & = \phi_2 \rho_2^{(t)},
\end{split}
\end{equation}

In the actual implementation of \texttt{PFDBNL}, we use a closed-form for $\tilde{W}_k,\tilde{A}_k$ as derived below: 

\[
\tilde{W}_k^{(t+1)} 
\;=\;
\frac{
  2\mu\,W_k^{(t+1)} 
  \;+\;
  \rho_2^{(t)}\,W^{(t)}
  \;-\;
  \beta_k^{(t)}
}{
  2\mu + \rho_2^{(t)}
}.
\]

\[
\tilde{A}_k^{(t+1)} 
\;=\;
\frac{
  2\mu\,A_k^{(t+1)} 
  \;+\;
  \rho_2^{(t)}\,A^{(t)}
  \;-\;
  \gamma_k^{(t)}
}{
  2\mu + \rho_2^{(t)}
}.
\]
For the full procedure, we have summarized it in the supplementary material \S~\ref{pclosed}. Similarly, in this work, we employ the L-BFGS method to solve the optimization problem. The procedure for \texttt{PFDBNL} is summarized in Algorithm~\ref{alg:PFDBNL}.
\begin{algorithm}[H]
\caption{Personalized Federated Dynamic Bayesian Network Learning}
\label{alg:PFDBNL}
\begin{algorithmic}[1]
\REQUIRE Initial parameters \( \rho_1, \rho_2, \alpha^{(1)}, \beta_1^{(1)}, \ldots, \beta_K^{(1)}, \gamma_1^{(1)}, \ldots, \gamma_K^{(1)} \); multiplicative factors \( \phi_1 , \phi_2 > 1 \); initial points \(\{W_k, A_k, \tilde{W}_k, \tilde{A}_k\}_{k = 1}^K, W, A \); participated clients at each round $j\leq K $
\FOR{$t = 1, 2, \ldots$}
    \STATE selected clients solve problem \eqref{p_local1} first and solve \eqref{p_local2} in parallel
    \STATE Central server collects \( \tilde{W}_{k_1}^{(t+1)}, \ldots, \tilde{W}_{k_j}^{(t+1)}, \tilde{A}_{k_1}^{(t+1)}, \ldots, \tilde{A}_{k_j}^{(t+1)} \) from all clients
    \STATE Central server solves problem \eqref{p_cen}
    \STATE Central server sends \( W^{(t+1)} \) and \( A^{(t+1)} \) to selected clients
    \STATE Central server updates ADMM parameters \( \alpha^{(t+1)}, \rho_1^{(t+1)}, \rho_2^{(t+1)} \) according to Eq.\eqref{para1}
    \STATE Each selected client updates \( \beta_k^{(t+1)}, \gamma_k^{(t+1)} \) according to Eq.\eqref{para1}
\ENDFOR
\end{algorithmic}
\end{algorithm}

\section{Experiments}

\begin{figure}[h!]
\centering
\includegraphics[width=0.8\textwidth]{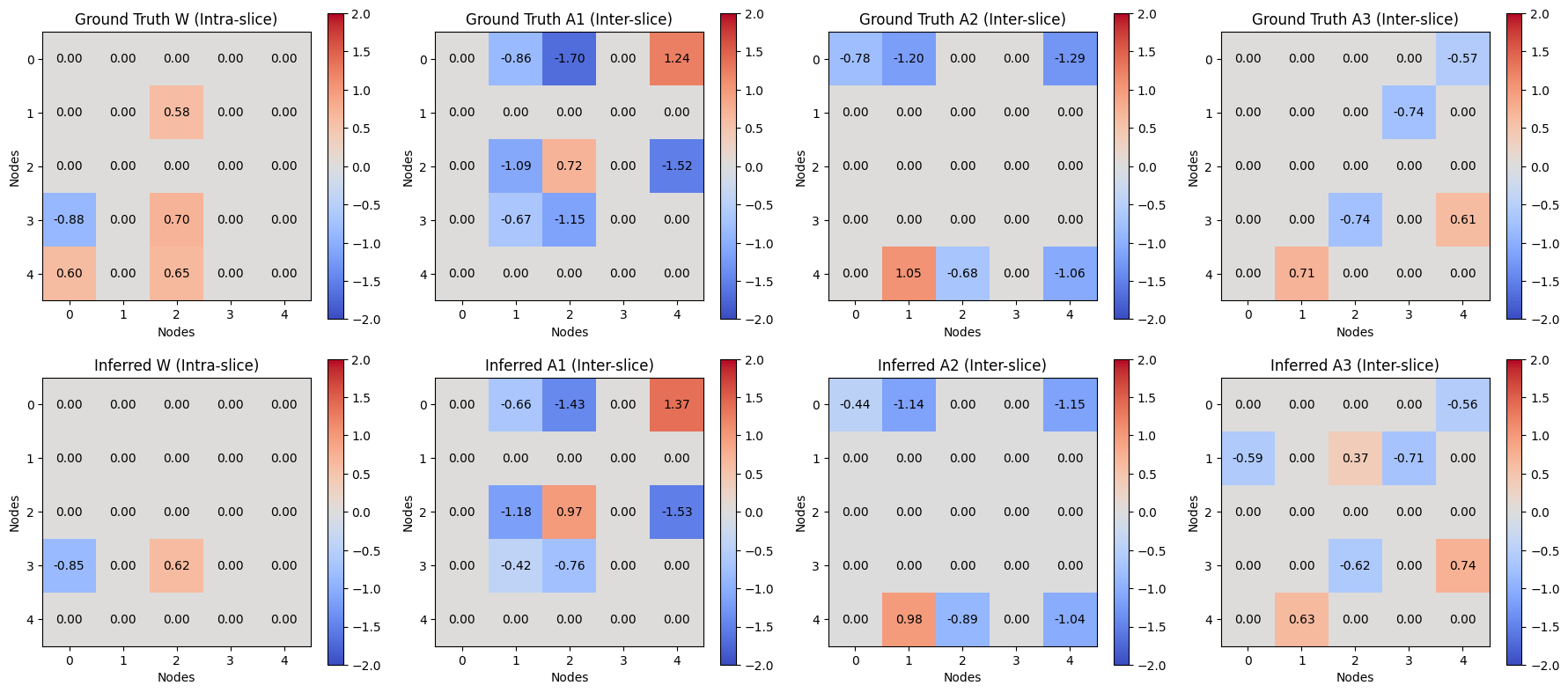}
\caption{An example result using \texttt{FDBNL} for Gaussian noise data with \(n = 500\) samples, \(d = 5\) variables, an autoregressive order \(p = 3\), and \(K = 10\) clients. All clients have same $W, A$. We set the thresholds \(\tau_w = \tau_a = 0.3\). Our algorithm recovers weights close to the ground truth.}
\label{example}
\end{figure}

In this section, we first evaluate the performance of \texttt{FDBNL} on simulated data generated from a linear SVAR structure \citep{gong2023causal}. We then compare it against three linear baseline methods using three evaluation metrics to demonstrate the effectiveness of our proposed method. Figure~\ref{example} provides an illustrative example of homogeneous time series data. Next, we examine the performance of \texttt{PFDBNL} on heterogeneous time series data generated by a linear SVAR structure across different clients and also analyze its performance under partial client participation by uniformly sampling a subset of clients. 

\paragraph{Benchmark Methods.}  
We compare our federated approach, described in Sec.~\ref{method} and referred to as \texttt{FDBNL}, with three other methods. The first baseline, denoted as \texttt{Ave}, computes the average of the weighted adjacency matrices estimated by DYNOTEARS \citep{pamfil2020dynotears} from each client’s local dataset, followed by thresholding to determine the edges. The second baseline, referred to as \texttt{Best}, selects the best graph from among those estimated by each client based on the lowest Structural Hamming Distance (SHD) to the ground truth. While this approach is impractical in real-world scenarios (since it assumes knowledge of the ground truth), it serves as a useful point of reference. For additional context, we also consider applying DYNOTEARS to the combined dataset from all clients, denoted as \texttt{Alldata}. Note that the final graphs produced by \texttt{Ave} may contain cycles for $W$, and we do not apply any post-processing steps to remove them, as doing so could degrade performance. Since there is no official source code for DYNOTEARS, we reimplement it using only the \texttt{numpy} and \texttt{scipy} packages in approximately a hundred lines of code. This simpler, self-contained implementation enhances both readability and reusability compared to existing versions on GitHub. For the personalized federated approach, referred to as \texttt{PFDBNL}, we compare its performance against \texttt{FDBNL} on heterogeneous time series data. Since no existing methods are specifically designed for learning heterogeneous DBNs, we focus on evaluating the improvements achieved by \texttt{PFDBNL} over \texttt{FDBNL} on a single example and demonstrate that \texttt{PFDBNL} produces reasonable results.

\paragraph{Evaluation Metrics.}  
We use three metrics to evaluate performance: Structural Hamming Distance (SHD), True Positive Rate (TPR), and False Discovery Rate (FDR). SHD measures the dissimilarity between the inferred graph and the ground truth by accounting for missing edges, extra edges, and incorrectly oriented edges \citep{tsamardinos2006max}. A lower SHD indicates a closer alignment with the ground truth. TPR (also known as sensitivity or recall) quantifies the proportion of true edges correctly identified. It is computed as the ratio of true positives to the sum of true positives and false negatives \citep{glymour2019review}. A higher TPR means the model successfully identifies more true edges. FDR measures the proportion of false positives among all predicted edges. It is defined as the ratio of false positives to the sum of false positives and true positives \citep{benjamini1995fdr}. A lower FDR indicates that most detected edges are correct. Together, these metrics provide a comprehensive assessment of the model’s accuracy in graph structure inference. Furthermore, for heterogeneous time series data, since we have \( K \) ground truth matrices \( W_k \) and \( A_{k_i} \), we evaluate performance separately for each client. Therefore, we report the mean SHD (mSHD), mean TPR (mTPR), and mean FDR (mFDR) across all clients. This is a reasonable evaluation metric, as it accounts for performance across all clients.

\subsection{Federated Dynamic Bayesian Network Learning}

\paragraph{Data Generation \& Settings.}  
For \texttt{FDBNL}, we generate data following the Structural Equation Model described in Eq.~\eqref{data}. This involves four steps: (1). Constructing weighted graphs \(G_W\) and \(G_A\); (2). Creating data matrices \(X\) and \(Y\) aligned with these graphs; (3). Partitioning the data among \(K\) clients as \(X_{client}\) and \(Y_{client}\); (4). Applying all algorithms to \(X_{client}\) and \(Y_{client}\) (or \(X\) and \(Y\)) and evaluating their performance. For details of steps (1) and (2), see \S\ref{sim}. We use Gaussian noise with a standard deviation of 1. The intra-slice DAG is an Erdős-Rényi (ER) graph with a mean degree of 4, and the inter-slice DAG is an ER graph with a mean out-degree of 1. Although this results in very sparse graphs at high \(d\), they remain connected under our settings by \citet{ErdosRenyi1960}. We set the base of the exponential decay of inter-slice weights to \(\eta=1.5\). For the hyperparameters \(\lambda_w\) and \(\lambda_a\), we generate heatmaps to systematically identify their optimal values. These values are documented in the Supplementary Materials (\S\ref{Hyper}). We set \(\phi_1 = 1.6\) and \(\phi_2 = 1.1\) with initial \(\rho_1=\rho_2=1\), and the initial Lagrange multipliers are zero. For DYNOTEARS (and thus \texttt{Ave} and \texttt{Best}), we follow the authors' recommended hyperparameters. To avoid confusion, we summarize the data shape used in \texttt{FDBNL}:
\begin{itemize}
    \item For \texttt{FDBNL}, \(X_{client}\) and \(Y_{client}\) are of size \(K \times n_k \times d\) and \(K \times n_k \times pd\), respectively, where \(K\) is the number of clients, \(n_k\) is the number of samples for $k$-th client, \(d\) is the dimensionality, and \(p\) is the autoregressive order.
\end{itemize}

\textbf{Experiment Settings.}  
We consider two types of experimental settings. First, we fix the number of clients \(K\) and increase the number of variables \(d\) while maintaining a consistent total sample size \(n\). This allows us to assess how well the methods scale with increasing dimensionality. Second, we fix the total sample size \(n \in \{256, 512\}\) and vary the number of clients \(K\) from 2 to 64. This setting evaluates adaptability and robustness as the data becomes more distributed.

\subsubsection{Varying Number of Variables}

In this section, we focus on DBN with \(n = 5d\) samples evenly distributed across \(K = 10\) clients for \(d = 10, 20\), and \(n = 6d\) for \(d \in \{5, 15\}\). Note that setting \(n = 5d\) or \(6d\) ensures that each client has an integer number of samples. We generate datasets for each of these cases. Typically, each client has very few samples, making this a challenging scenario.

\begin{figure}[H]
\centering
\includegraphics[width=1\textwidth]{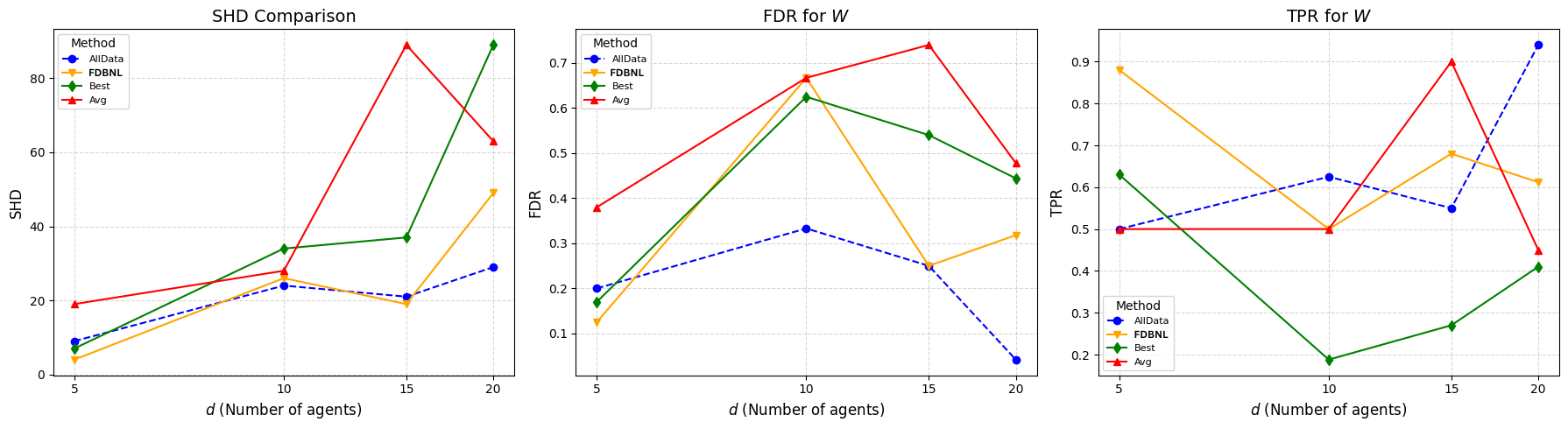}
\caption{Structure learning results for \(W\) in a DBN with Gaussian noise for \(d = 5, 10, 15, 20\) variables, an autoregressive order \(p = 1\), and \(K = 10\) clients. Each metric value indicates the mean performance across 10 different simulated datasets}
\label{vary_d}
\end{figure}

Figure \ref{vary_d} shows the results for the inferred \(W\). The corresponding results for \(A\) are provided in \S\ref{result_a}. Across all tested values of \(d\), \texttt{FDBNL} consistently achieves the lowest SHD compared to \texttt{Ave} and \texttt{Best}, and even outperforms \texttt{Alldata} for $d = 5, 15$. This advantage may be due to the need for more refined hyperparameter tuning in \texttt{Alldata}. In addition to a lower SHD, \texttt{FDBNL} achieves a relatively high TPR, close to that of \texttt{Alldata} and higher than that of \texttt{Best}, indicating that \texttt{FDBNL} accurately identifies most true edges. Moreover, \texttt{FDBNL} attains a lower FDR than \texttt{Ave} and \texttt{Best}, emphasizing its superior reliability in edge identification under this challenging setup.

\subsubsection{Varying Number of Clients}

We now examine scenarios where a fixed total number of samples is distributed among varying numbers of clients. For \(d \in \{10, 20\}\), we generate \(n = 512\) samples with \(p = 1\). These are evenly allocated across \(K \in \{2, 4, 8, 16, 32, 64\}\) clients. For \(\lambda_w\) and \(\lambda_a\), we follow the recommendations of \cite{pamfil2020dynotears} for DYNOTEARS, \texttt{Ave}, and \texttt{Best}. For \texttt{FDBNL}, we select the best regularization parameters from \([0.05, 0.5]\) in increments of 0.05 by minimizing SHD.

\begin{figure}[H]
\centering
\includegraphics[width=\textwidth]{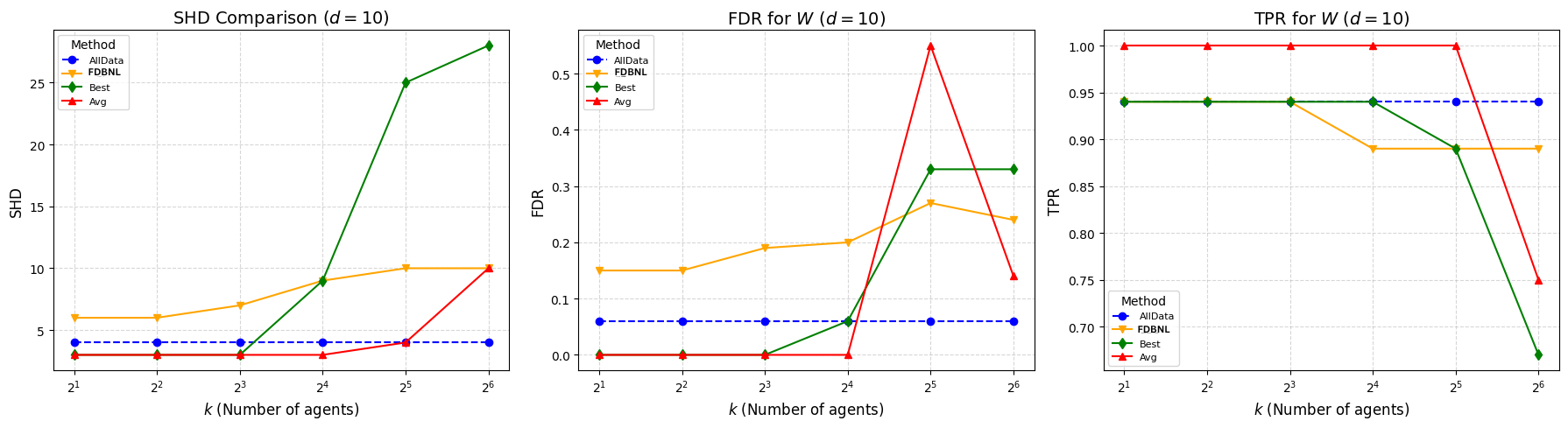}
\caption{Structure learning for \(W\) of a DBN with Gaussian noise for \(d = 10\) variables, \(p = 1\), and varying numbers of clients. There are \(n = 512\) total samples, distributed evenly across \(K \in \{2, 4, 8, 16, 32, 64\}\). Each metric value indicates the mean performance across 10 different simulated datasets}
\label{10d_512client}
\end{figure}

\begin{figure}[H]
\centering
\includegraphics[width=\textwidth]{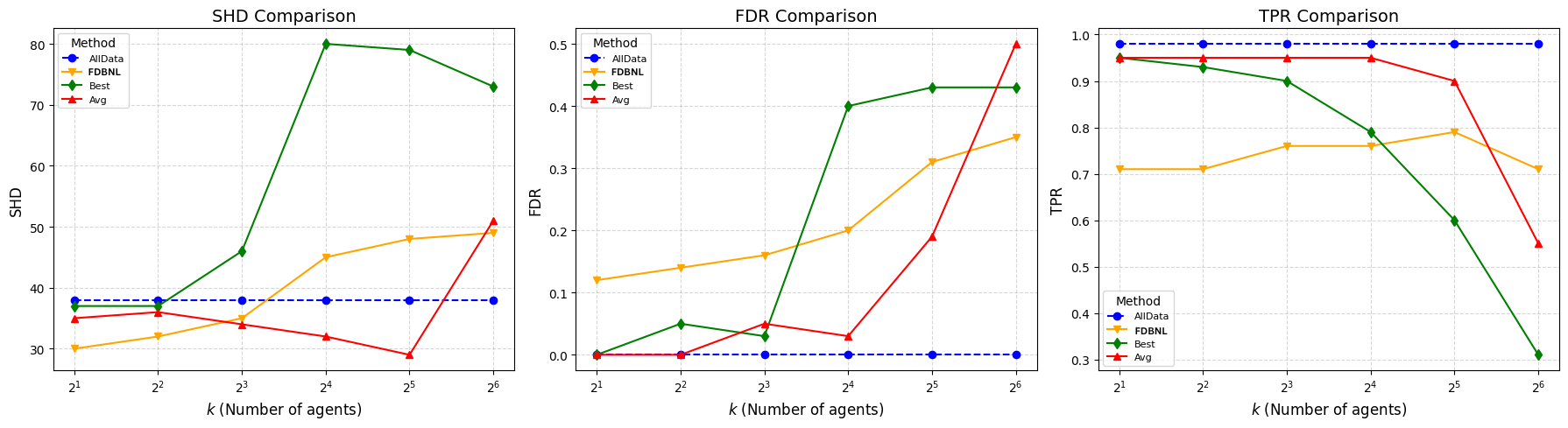}
\caption{Structure learning for \(W\) of a DBN with Gaussian noise for \(d = 20\) variables, \(p = 1\), and varying numbers of clients. There are \(n = 512\) total samples, distributed evenly across \(K \in \{2, 4, 8, 16, 32, 64\}\). Each metric value indicates the mean performance across 10 different simulated datasets}
\label{20d_512client}
\end{figure}

Figures \ref{10d_512client} and \ref{20d_512client} show the results for \(W\) with \(d = 10\) and \(d = 20\). The results for \(A\) are in \S\ref{result_a}. As \(K\) increases, the TPRs of \texttt{Ave} and \texttt{Best} drop sharply, resulting in high SHDs. Although \texttt{FDBNL}'s TPR also decreases as \(K\) grows, it remains significantly higher than those of the other baselines. For example, with \(d = 20\) and \(K = 64\), \texttt{FDBNL} achieves a TPR of 0.7, while \texttt{Ave} and \texttt{Best} achieve only 0.5 and 0.3, respectively. This highlights the importance of information exchange in the optimization process, enabling \texttt{FDBNL} to learn a more accurate DBN structure even in highly distributed scenarios.

Next, we consider the case of \(n = 256\) samples. The corresponding \(W\)-based results are shown below, and \(A\)-based results are in \S\ref{result_a}.

\begin{figure}[H]
\centering
\includegraphics[width=\textwidth]{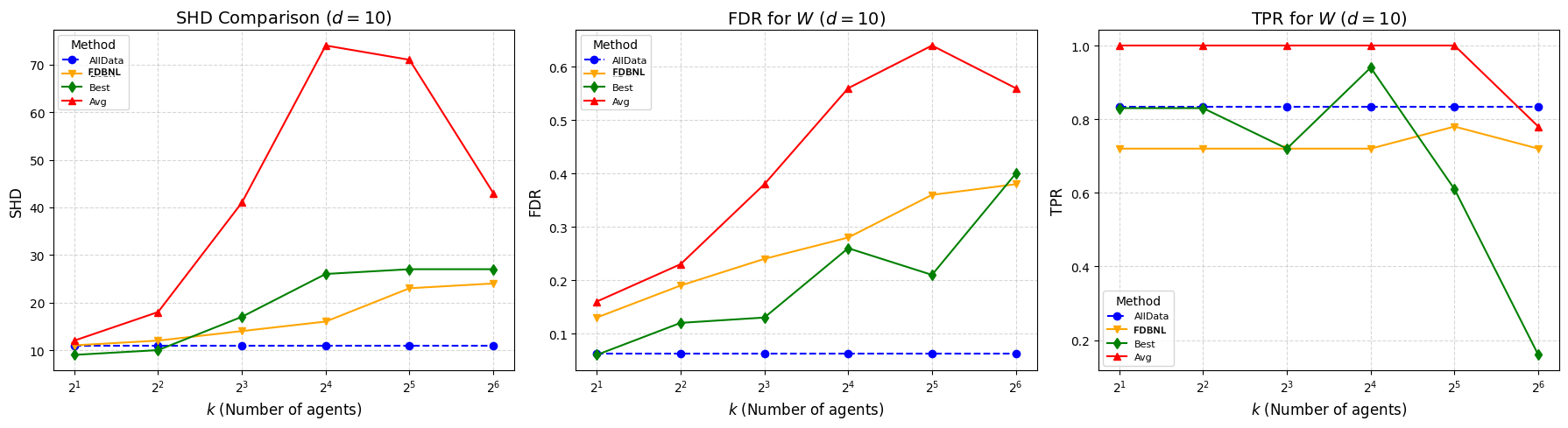}
\caption{Structure learning for \(W\) of a DBN with Gaussian noise for \(d = 10\) variables, \(p = 1\), and varying numbers of clients. There are \(n = 256\) total samples, distributed evenly across \(K \in \{2, 4, 8, 16, 32, 64\}\). Each metric value indicates the mean performance across 10 different simulated datasets} 
\label{10d_256client}
\end{figure}

\begin{figure}[H]
\centering
\includegraphics[width=\textwidth]{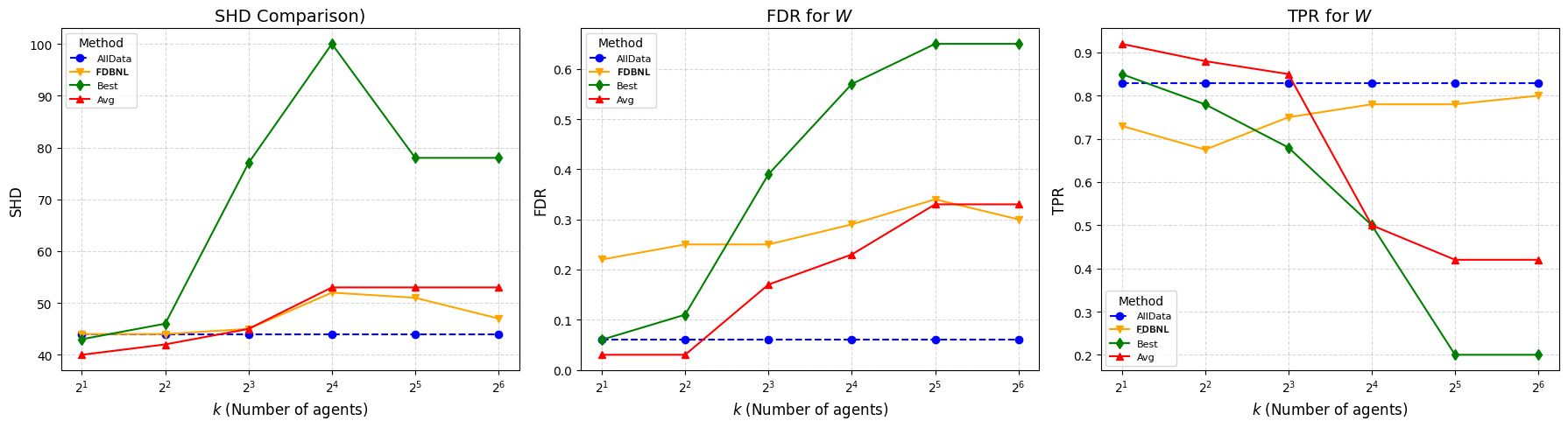}
\caption{Structure learning for \(W\) of a DBN with Gaussian noise for \(d = 20\) variables, \(p = 1\), and varying numbers of clients. There are \(n = 256\) total samples, distributed evenly across \(K \in \{2, 4, 8, 16, 32, 64\}\). Each metric value indicates the mean performance across 10 different simulated datasets} 
\label{20d_256client}
\end{figure}

From Figures \ref{10d_256client} and \ref{20d_256client}, \texttt{FDBNL} consistently attains the lowest SHD compared with \texttt{Best} and \texttt{Ave}. For \(d = 20\), the TPRs of \texttt{Ave} and \texttt{Best} continue to decline as \(K\) increases, whereas \texttt{FDBNL}'s TPR remains higher for all \(K \geq 8\). At extremely high distributions, such as \(K = 64\) with \(d = 20\), each client has only 4 samples, leading to non-convergence for \texttt{Ave} and \texttt{Best}. Thus, we compare metrics at \(K = 32\) for a fair assessment. For \(d = 10\), while \texttt{Ave} achieves the highest TPR among all methods for all \(K\), its FDR is much larger than that of the other methods. This suggests that the thresholding strategy used by \texttt{Ave} may lead to a high number of false positives. Nonetheless, \texttt{FDBNL}'s TPR for \(d = 10\) remains close to that of \texttt{Alldata} as \(K\) grows.

In some cases, when \(K\) or \(d\) is small, the other baselines may outperform \texttt{FDBNL}. For example, with \(d = 10, n = 512\), \texttt{Ave} and \texttt{Best} achieve better SHDs than \texttt{FDBNL} when \(K \leq 16\). This is expected since fewer clients and lower dimensionality provide each client with enough samples for accurate independent structure learning. However, as complexity grows (e.g., \(d = 20\)), \texttt{FDBNL} outperforms the baselines by maintaining a lower FDR and effectively leveraging distributed information. These findings highlight the importance of both method selection and hyperparameter tuning, as well as potential directions for future research in improving the efficiency of information exchange mechanisms for DBN learning.

\subsection{Personalized Federated Dynamic Bayesian Network Learning}
In this section, we first compare \texttt{PFDBNL} to \texttt{FDBNL}. For this example, we generate six pairs of \( (W, A) \) for each client. Then, we simulate 10 datasets with \( n_k = 30 \) for the $k$-th client. We compute the mean mSHD, mTPR, and mFDR after applying \texttt{PFDBNL} and \texttt{FDBNL} to each of these datasets.  

From Table~\ref{compare_two}, We observe that the mSHD of \( W \) and \( A \) in \texttt{PFDBNL} is half that of \( W \) and \( A \) in \texttt{FDBNL}. Furthermore, the mTPR of \texttt{PFDBNL} is twice as high as that of \texttt{FDBNL}. For mFDR, we observe that \texttt{PFDBNL}'s value is slightly lower than \texttt{FDBNL}'s, but the mFDR decreases by a factor of two. Hence, \texttt{PFDBNL} significantly improves performance across all evaluation metrics compared to \texttt{FDBNL}. \textbf{To the best of our knowledge, no existing method is capable of learning Dynamic Bayesian Networks (DBNs) from heterogeneous time-series data}. Given this gap, \texttt{PFDBNL} represents the first attempt to address this challenge and demonstrates its effectiveness in improving accuracy.

\begin{table}[h!]
\label{compare_two}
\centering
\begin{tabular}{lccc}
\hline
Metric &  & \texttt{PFDBNL} & \texttt{FDBNL} \\ \hline
mSHD    & $W$ &  6.2   &  10.2   \\
       & $A$ &  5.4    &  12.7   \\ \hline
mTPR    & $W$ &  0.64  &  0.35  \\
       & $A$ &  0.51   &  0.24  \\ \hline
mFDR    & $W$ &  0.55  &  0.60  \\
       & $A$ &  0.19   &  0.33 \\ \hline
\end{tabular}
\caption{Comparison between \texttt{FDBNL} and \texttt{PFDBNL} with $d = 5$, $K = 6$, $n_k = 30$ for $k \in K$. The expected degree of a node is 4 (both direction) with $p = 1$. Each metric value indicates the mean performance across 10 different simulated datasets}
\end{table}

\paragraph{Data Generation \& Settings.}  
For \texttt{PFDBNL}, we generate data following the Structural Equation Model described in Eq.~\eqref{data}. This involves three steps: (1) Constructing weighted graphs \( G_{W_k} \) and \( G_{A_{k_i}} \) for each client. (2) Creating data matrices \( X_{\text{client}} \) and \( Y_{\text{client}} \) aligned with these graphs for each client. (3) Applying \texttt{PFDBNL} and evaluating its performance. Steps (1) and (2) follow the same data generation procedure as \texttt{FDBNL}, but partitioning of the dataset is not required. We use Gaussian noise with a standard deviation of 1 and set the base of the exponential decay of inter-slice weights to \( \eta = 1.5 \). For the hyperparameters \( \lambda_W \), \( \lambda_A \), and \( \mu \), we found that setting \( \lambda_W = \lambda_A = 0.1 \) works well for \( d = \{5,10\} \), while \( \lambda_W = \lambda_A = 0.01 \) is effective for \( d = \{15,20\} \) with \( \mu = 0.1 \). We set \( \phi_1 = 1.6 \) and \( \phi_2 = 1.1 \), with initial values \( \rho_1 = \rho_2 = 1 \). The initial Lagrange multipliers are set to zero. The data shape is the same as in the \texttt{FDBNL} experiment, but each client has a different \( W_k, A_{k_i} \).

\textbf{Experiment Settings.}  
We consider the same two types of experimental settings as in the \texttt{FDBNL} experiment. First, we fix the number of clients \( K \) and increase the number of variables \( d \) while maintaining a consistent sample size \( n \). This allows us to assess how well the methods scale with increasing dimensionality. Second, we fix the total sample size at \( n = 512 \) and vary the number of clients \( K \) from 2 to 32. This setting evaluates the adaptability and robustness of the methods as the data becomes more distributed.

\subsubsection{Varying Number of Variables}
In this section, we focus on learning of \( W_k \) and \( A_{k_i} \) with each \(n_k = Kd\) samples for \(d \in \{5, 10, 15, 20\}\) across \(K = 6\) clients. Similarly to the example, we firstly generate $W_k, A_{k_1}$ by using $p = 1$ for all clients. Then we generate the dataset with $n_k$ sample sizes for $k$-th client. Furthermore, we exam our \texttt{PFDBNL} under two different connectivity level (one for low connectivity and one for high connectivity). We do not focused on mixed case in this study and leave it for future research direction as discussed in \S\ref{dis}. For low connectivity, we set the expectation of degree of a node below half of number of variable as $\lfloor \frac{d}{2} \rfloor - \frac{d}{5} $. For high connective, we set expectation of degree of a node as $d - \frac{d}{5}$.

\begin{table}[H]
\centering

\begin{tabular}{lcccccc}
\hline
Metric & & 5 & 10 & 15 & 20 \\ \hline
mSHD   & $W$ & 4.4  & 16.8  & 44.7  & 79.2  \\ 
       & $A$ & 2.8  & 10.9  & 29.6  & 32.2  \\ \hline
mTPR   & $W$ & 0.61 & 0.49  & 0.55  & 0.53  \\ 
       & $A$ & 0.81 & 0.53  & 0.31  & 0.31  \\ \hline
mFDR   & $W$ & 0.72 & 0.67  & 0.51  & 0.53  \\ 
       & $A$ & 0.35 & 0.47  & 0.23  & 0.34  \\ \hline
\end{tabular}
\caption{Comparison of metrics (mSHD, mTPR, mFDR) across dimensions $d = 5, 10, 15, 20$ for low connectivity. Each metric value indicates the mean performance across 10 different simulated datasets}
\label{all_sp}
\end{table}

\begin{table}[H]
\centering

\begin{tabular}{lcccccc}
\hline
Metric & & 5 & 10 & 15 & 20 \\ \hline
mSHD   & $W$ & 6.2  & 13.6  & 52.0  & 110.0  \\ 
       & $A$ & 5.4  & 8.3  & 41.0  & 31.0   \\ \hline
mTPR   & $W$ & 0.64 & 0.64  & 0.71  & 0.76   \\ 
       & $A$ & 0.51 & 0.52  & 0.31  & 0.27   \\ \hline
mFDR   & $W$ & 0.55 & 0.54 & 0.44  & 0.59   \\ 
       & $A$ & 0.19 & 0.21  & 0.36  & 0.19   \\ \hline
\end{tabular}
\caption{Comparison of metrics (mSHD, mTPR, mFDR) across dimensions $d = 5, 10, 15, 20$ for high connectivity. Each metric value indicates the mean performance across 10 different simulated datasets}
\label{all_de}
\end{table}
 
From Tables~\ref{all_sp} and \ref{all_de}, we observe that the mSHD for the low-connectivity case is much lower than in the high-connectivity case for \( W \). However, the mSHD remains similar across all cases for \( A \). For mTPR, we observe that performance in the high-connectivity scenario is significantly better than in the low-connectivity scenario. More importantly, while the mTPR for high-connectivity does not decrease with an increasing number of variables for \( W \), it does decrease for \( A \). Additionally, the mTPR for \( A \) remains similar for \( d \in \{10, 15, 20\} \), but in the low-connectivity case, it is 0.81, which is significantly higher than 0.51 in the high-connectivity case. This is not surprising, as identifying relationships in sparse networks is more challenging. For mFDR of \( A \), we observe that it remains around 0.2 for \( d \in \{5, 10, 20\} \) in the high-connectivity case. However, for the low-connectivity case, the mFDR increases to approximately 0.3. For \( W \), the mFDR remains stable at around 0.54 in the high-connectivity scenario, which is significantly lower than in the low-connectivity case when \( d \in \{5, 10\} \). Overall, the mFDR in the high-connectivity case shows better results.

We further explore the partial participation case under \( d = 10 \). Since we only have six clients, we set the number of participating clients to \( j = 1, 2, 3, 4, 5 \), while keeping all other settings unchanged. As shown in Table~\ref{partial}, the results remain stable regardless of the number of participating clients. The mSHD is approximately 13 to 14 for \( W \) and around 8 for \( A \). The mTPR remains about 0.60 for \( W \) and 0.56 for \( A \). The mFDR follows a similar pattern.

\begin{table}[H]
\centering
\begin{tabular}{lcccccc}
\hline
Metric & & $ j = 1$  &$ j = 2$  & $j = 3$& $j = 4$ & $j = 5$ \\ \hline
mSHD   & $W$ & 14.2 & 13.6 & 14.3 & 14.9 & 14.5\\ 
       & $A$ & 8.1 & 8.3 & 8.7 & 7.7 & 8.5\\ \hline
mTPR   & $W$ & 0.59 & 0.61 & 0.60 & 0.58 & 0.60\\  
       & $A$ & 0.56 & 0.55 & 0.55 & 0.58 & 0.56\\  \hline
mFDR   & $W$ & 0.50 & 0.49 & 0.53 & 0.55 & 0.54\\ 
       & $A$ & 0.22 & 0.21 & 0.23 & 0.23 & 0.23\\  \hline
\end{tabular}
\caption{Partial participating case for $d = 10$. Each metric value indicates the mean performance across 10 different simulated datasets}
\label{partial}
\end{table}

\subsubsection{Varying Number of Clients}

We now examine scenarios where a fixed total number of samples is distributed among varying numbers of clients. For \( d \in \{10, 20\} \), we generate \( n = 512 \) samples in total. These samples are evenly distributed across \( K \in \{2, 4, 8, 16, 32\} \) clients, meaning that each client receives \( n_k = \{256, 128, 64, 32, 16\} \) samples, respectively. The mean degree of the intra-slice DAG is 4, counting edges in both directions. This may be a challenging setting because we need to learn different DBNs for 32 clients with only 16 samples per client.
\begin{table}[H]
\centering
\begin{tabular}{lcccccc}
\hline
Metric & & $K = 2$  & $K = 4$& $K = 8$ & $K = 16$ & $K = 32$\\ \hline
mSHD   & $W$ & 11.3 & 20.6 & 17.4 & 18.2 & 19.4  \\ 
       & $A$ & 6.7 & 10.5  & 10.7 & 10.8 & 14.8   \\ \hline
mTPR   & $W$ & 0.68 & 0.44 & 0.51 & 0.51 & 0.49 \\ 
       & $A$ & 0.63 & 0.52 & 0.48 & 0.48 & 0.36 \\ \hline
mFDR   & $W$ & 0.49 & 0.62 & 0.54  & 0.54 & 0.59 \\ 
       & $A$ & 0.17 & 0.31 & 0.23  & 0.28 & 0.37 \\ \hline
\end{tabular}
\caption{Comparison of metrics (mSHD, mTPR, mFDR) across dimensions $K = 2, 4, 8, 16, 32$ for $d = 10$. Each metric value indicates the mean performance across 10 different simulated datasets}
\label{VC_10}
\end{table}

\begin{table}[H]
\centering

\begin{tabular}{lcccccc}
\hline
Metric & & $K = 2$  & $K = 4$& $K = 8$ & $K = 16$ & $K = 32$\\ \hline
mSHD   & $W$ & 56.2 & 57.5 & 54.3 & 52.1 & 56.1  \\ 
       & $A$ & 33.2 & 36.1 & 40.8 & 55.6 & 56.3 \\ \hline
mTPR   & $W$ & 0.45 & 0.46 & 0.46 & 0.34 & 0.32  \\ 
       & $A$ & 0.38 & 0.31 & 0.33 & 0.25 & 0.17 \\ \hline
mFDR   & $W$ & 0.40 & 0.45 & 0.37 & 0.49 & 0.51 \\ 
       & $A$ & 0.18 & 0.16 & 0.16 & 0.21 & 0.27 \\ \hline
\end{tabular}
\caption{Comparison of metrics (mSHD, mTPR, mFDR) across dimensions $K = 2, 4, 8, 16, 32$ for $d = 20$. Each metric value indicates the mean performance across 10 different simulated datasets}
\label{VC_20}
\end{table}

From Tables~\ref{VC_10} and \ref{VC_20}, we observe that \texttt{PFDBNL} consistently achieves mSHD values around 10–20 for \( W \) and approximately 10 for \( A \) when \( d=10 \). Notably, the mTPR of \( W \) remains relatively stable as \( K \) increases. For instance, at \( K=32 \), \( W \) and \( A \) obtain mTPRs of 0.49 and 0.36, respectively. This result is encouraging, even in scenarios where the number of clients (\( K \)) exceeds the number of samples per client. Meanwhile, the mFDR remains around 0.2–0.3 for \( A \) but roughly 0.5 for \( W \), suggesting that \( A \) is more conservative, with fewer false discoveries.  

For \( d=20 \), although \( W \) stabilizes at an mSHD of around 55, the performance of \( A \) deteriorates from an mSHD of approximately 30 to 50 as \( K \) increases, indicating greater sensitivity to high-dimensional settings. A similar pattern is observed in the mTPR: \( W \) achieves approximately 0.45 when \( K \in \{2,4,8\} \) but declines to about 0.3 for larger \( K \). This drop is expected, as the sample size per client decreases while the number of parameters in the ground-truth matrix remains large. Overall, however, \texttt{PFDBNL} maintains stable and reasonable performance across different values of \( K \), including the challenging scenario of \( d=20, K=32, n_k=16 \).

\section{Applications}
\subsection{DREAM4}

Building on the DREAM4 Challenge \citep{marbach2009generating,stolovitzky2007dialogue,stolovitzky2009lessons},
our paper focuses on the time-series track of the \textit{InSilico\_Size100} subchallenge. In this problem, the DREAM4 gene expression data are used to infer gene regulatory networks. The \textit{InSilico\_Size100} subchallenge dataset contains 5 independent datasets, each consisting of 10 time-series for 100 genes, measured over 21 time steps. We assume that each dataset corresponds to data collected from different hospitals or research centers, thereby motivating our federated approach. More information and data about DREAM4 are available at \url{https://gnw.sourceforge.net/resources/DREAM4%20in%20silico%20challenge.pdf}

Let \( X^g_{t,r} \) denote the expression level of gene \( g \) at time \( t \in \{0,1,2,\ldots,20\} \) in replicate \( r \in \{0,1,2,\ldots,R\} \). Note that \( R \) depends on the dataset used—for instance, if only one dataset is used, then \( R = 10 \). Consequently, \( X_{t,r} \in \mathbb{R}^{100} \) and \( X_t \in \mathbb{R}^{R \times 100} \). In this experiment, we set \( p = 1 \), aligning with the VAR method proposed in the DREAM4 Challenge by \citet{lu2021causal}. The federated setting of our \texttt{FDBNL} approach includes \( K = 5 \) clients, each with \( R = 2 \) replicates. Thus, each client contains a time-series dataset for 100 genes over 42 time steps. We found that small regularization parameters, \( \lambda_W = \lambda_A = 0.0025 \), work well for all DREAM4 datasets. For our \texttt{PFDBNL} approach, we assume each dataset to be a client. Consequently, each client has 200 time steps with \( d = 100 \) and \( K = 5 \). We used regularization parameters \( \lambda_W = \lambda_A = 0.001 \) and \( \mu = 0.1 \).

\citet{lu2021causal} evaluated various methods for learning these networks, including approaches based on Mutual Information (MI), Granger causality, dynamical systems, Decision Trees, Gaussian Processes (GPs), and Dynamic Bayesian Networks (DBNs). Notably, we did not compare DYNOTEARS in this study, as its source code is not publicly available, and our implemented version did not achieve optimal results. The comparisons were made using AUPR (Area Under the Precision-Recall Curve) and AUROC (Area Under the Receiver Operating Characteristic Curve) across the five datasets. 

From Table~\ref{dream4_comp}, the GP-based method outperforms all others, achieving the highest mean AUPR (0.208) and AUROC (0.728). In terms of AUPR, our \texttt{FDBNL} approach outperforms the TSNI (ODE-based) method in its non-federated version. Furthermore, the AUPR of \texttt{FDBNL} is comparable to those of Ebdnet (DBN-based), GCCA (VAR-based), and ARACNE (MI-based) approaches. Importantly, the mean AUROC of \texttt{FDBNL} surpasses those of GCCA (VAR-based), ARACNE (MI-based), and TSNI (ODE-based) methods, and is close to those of Ebdnet and VBSSMa (DBN-based). We also directly applied \texttt{FDBNL}, marked as \texttt{FDBNL-d}, to all datasets using the same settings as \texttt{PFDBNL}. From the table, we can see that if we directly use \texttt{FDBNL}, the results are not satisfactory. Therefore, we partitioned each dataset to ensure i.i.d. conditions for the use of \texttt{FDBNL}. We then tested \texttt{PFDBNL}, which performs well in heterogeneous situations. For \texttt{PFDBNL}, the AUPR is 0.034, which outperforms the TSNI (ODE-based) method in its non-personalized version. Additionally, the AUPR of \texttt{FDBNL} is only slightly greater than that of \texttt{PFDBNL}. Moreover, \texttt{PFDBNL} is also comparable to Ebdnet (DBN-based), GCCA (VAR-based), and ARACNE (MI-based) approaches. More importantly, the mean AUROC of \texttt{PFDBNL} surpasses that of ARACNE (MI-based) and is close to those of TSNI (ODE-based) and GCCA (VAR-based) methods. 

Overall, \texttt{FDBNL} and \texttt{PFDBNL} \textbf{are comparable or even sometimes superior to other non-federated benchmarks}. However, the GP-based method still achieves superior results in both AUPR and AUROC. One possible reason for this superiority is that the GP-based approach may more effectively capture nonlinear relationships and complex temporal dynamics, which can be challenging in a federated setting where data distributions and conditions vary across sources. Unsurprisingly, the GP-based method outperforms all approaches because Gaussian Processes naturally handle nonlinear dynamics and are more adaptive to varying, complex relationships. In contrast, the \texttt{FDBNL} and \texttt{PFDBNL} models, although well-suited for distributed data, may impose stronger parametric assumptions or face computational constraints that limit their ability to capture subtle nonlinear interactions. Thus, we believe GP-based methods could be extended to a distributed approach for structure learning, making them more applicable to real-world problems. We discuss this further in Section~\ref{dis}.

\begin{table}[H]
\centering
\begin{tabular}{lccc}
\hline
\textbf{Method} & \textbf{Type} & \textbf{AUPR} & \textbf{AUROC} \\ \hline
FDBNL    & F-DBN & 0.040 & 0.600 \\ \hline
PFDBNL    & PF-DBN & 0.034 & 0.561\\ \hline
FDBNL-d    & F-DBN & 0.023 & 0.512\\ \hline
Ebdnet \citep{rau2010empirical}  & DBN     & 0.043 & 0.640 \\ \hline
VBSSMa \citep{penfold2011gene}   & DBN     & 0.086 & 0.620 \\ \hline
CSId \citep{penfold2015csi}  & GP      & 0.208 & 0.728 \\ \hline
GCCA \citep{penfold2011gene}   & VAR     & 0.050 & 0.584 \\ \hline
TSNI \citep{penfold2011gene}    & ODE     & 0.026 & 0.566 \\ \hline
ARACNE \citep{margolin2006aracne}  & MI      & 0.046 & 0.558 \\ \hline
\end{tabular}
\caption{Comparison of mean AUPR and mean AUROC scores on the DREAM4 dataset.}
\label{dream4_comp}
\end{table}

\subsection{Functional Magnetic Resonance Imaging (FMRI)}

In this experiment, we apply the proposed learning methods to estimate connections in the human brain using simulated blood oxygenation level-dependent (BOLD) imaging data \citep{smith2011network}. The dataset consists of 28 independent datasets with the number of observed variables \(d \in \{5, 10, 15\}\). Each dataset contains 50 subjects (i.e., 50 ground-truth networks) with 200 time steps. To conduct the experiments, we use simulated time series measurements corresponding to five different human subjects for each \(d\) and compute the Average AUROC using the \texttt{sklearn} package. 

For the \texttt{FDBNL} setting, we partition the 200 time steps among five clients (\( K = 5 \)). Detailed information and descriptions of the data are available at \url{https://www.fmrib.ox.ac.uk/datasets/netsim/index.html}. In our experiments, we evaluate the proposed method for \( d \in \{5, 10, 15\} \). For \( d = 5 \), we use the 3rd, 6th, 9th, 12th, and 15th subjects from \texttt{Sim-1.mat}. For \( d = 10 \), we use the 2nd, 4th, 6th, 8th, and 10th subjects from \texttt{Sim-2.mat}. For \( d = 15 \), we use the 1st, 3rd, 5th, 7th, and 9th subjects from \texttt{Sim-3.mat}. For the \texttt{PFDBNL} setting, we do not apply any partitioning; instead, we consider each subject as a client and we aaply our \texttt{PFDBNL} to 5 subjects. Thus, for \( d \in \{5, 10, 15\} \), we have five clients, each containing 200 time steps. For \( d = 5 \), we use the 2nd, 4th, 6th, 8th, and 10th subjects from \texttt{Sim-1.mat}. For \( d = \{10, 15\} \), we use the 1st, 3rd, 5th, 7th, and 9th subjects from \texttt{Sim-2.mat} and \texttt{Sim-3.mat}. To avoid any confusion, the difference between the two methods is as follows: (1) \texttt{FDBNL} is applied to each subject individually, where each subject is partitioned into five clients. As a result, we need to run five repetitions. (2) \texttt{PFDBNL} is directly applied to the five subjects without partitioning. Further details of this experiment are provided in the supplementary materials (see \S\ref{app_FMRI}).

We compare our method to the Economy Statistical Recurrent Units (eSRU) model proposed by \citet{Khanna2019EconomySR} for inferring Granger causality, as well as existing methods based on a Multilayer Perceptron (MLP), a Long Short-Term Memory (LSTM) network \citep{Tank2022}, and a Convolutional Neural Network (CNN)-based model, the Temporal Causal Discovery Framework (TCDF) \citep{make1010019}, on multivariate time series data for \( d = 15 \), as examined by \citet{Khanna2019EconomySR}. As shown in Table~\ref{perf:d15}, our proposed \texttt{FDBNL} achieves an AUROC of 0.74, outperforming the LSTM-based approach and approaching the performance of the CNN-based TCDF method. Although \texttt{FDBNL} does not surpass the MLP-based and eSRU-based methods, it is noteworthy that its \textbf{federated version outperforms or closely matches several established non-federated benchmarks}. Similarly, our proposed \texttt{PFDBNL} also achieves an AUROC of 0.74, demonstrating superior performance over the LSTM-based approach and approaching the CNN-based TCDF method. Notably, \texttt{PFDBNL} can be directly applied to multiple subjects, significantly reducing the computational cost associated with data preparation. While \texttt{PFDBNL} does not exceed the MLP-based and eSRU-based methods, its personalized version consistently outperforms or closely aligns with several established non-personalized benchmarks. This outcome is expected, as the MLP and eSRU methods rely on deep architectures that excel at modeling complex structural dependencies.  

Importantly, our \texttt{FDBNL} and \texttt{PFDBNL} methods offer a new perspective on this problem by ensuring data security through their federated and personalized approach. This capability is particularly crucial in scenarios involving sensitive or distributed datasets, as it enables effective analysis without compromising privacy or data integrity. We further elaborate on this aspect in Section~\ref{dis}.

\begin{table}[H]
\centering
\begin{tabular}{lc}
\hline
\textbf{Method} & \textbf{Averaged AUROC} \\ \hline
MLP \citep{Tank2022}   & 0.81$\pm$0.04 \\ 
LSTM \citep{Tank2022}  & 0.70$\pm$0.03 \\ 
TCDF \citep{make1010019}  & 0.75$\pm$0.04 \\ 
eSRU \citep{Khanna2019EconomySR} & 0.84$\pm$0.03 \\ 
FDBNL & \textbf{0.74}$\pm$0.04 \\ 
PFDBNL & \textbf{0.74}$\pm$0.05 \\ \hline
\end{tabular}
\caption{Mean AUROC comparison of different methods for \(d = 15\).}
\label{perf:d15}
\end{table}

\section{Discussion}
\label{dis}
We proposed a federated framework for learning Dynamic Bayesian Networks (DBNs) on homogeneous time series data and a personalized federated framework for heterogeneous time series data. Specifically, we designed a federated DBN learning method using ADMM, where only model parameters are exchanged during optimization. Our experimental results demonstrate that this approach outperforms other state-of-the-art methods, particularly in scenarios with numerous clients, each possessing a small sample size—a common situation in federated learning that motivates client collaboration. Additionally, we developed a personalized federated DBN learning method by incorporating a proximal operator and utilizing ADMM. Our experimental results indicate that the personalized approach improves accuracy while maintaining robust performance even with small sample sizes and a large number of clients. Furthermore, we demonstrate that partial client participation does not lead to a loss of accuracy. Below, we address some limitations of our approach and suggest potential directions for future research.

\paragraph{Assumptions.}
We assume that the DBN structure, represented by \(W, A\), remains constant over time and is the same for all time series in the dataset. Relaxing this assumption could be useful in various ways, such as allowing the structure to change smoothly over time \citep{LeTV2009}. Another direction for future work is to investigate the behavior of the algorithm on non-stationary or cointegrated time series \citep{malinsky19a} or in scenarios with confounding variables \citep{Huang2015ITD}.

\paragraph{Federated Learning.}
In our \texttt{FDBNL}, the ADMM-based approach to federated learning relies on a stateful setup, requiring each participating client to be involved in every round of communication and optimization. Even though we attempt to implement partial client participation for \texttt{PFDBNL}, applying the same strategy is meaningless since it only affects which clients join, without altering the fundamental process. This “always-on” requirement can be burdensome in real-world scenarios. For instance, in large-scale deployments, clients such as mobile devices or IoT sensors may experience intermittent connectivity, limited power, or varying levels of availability. Ensuring that all such devices participate consistently and synchronously in every round is often impractical and can result in significant performance bottlenecks. An important future direction is to apply other federated optimization techniques and explore asynchronous techniques, which would enable stateless clients and facilitate cross-device learning. Additionally, the ADMM procedure involves sharing model parameters with a central server, raising concerns about potential privacy risks. Research has shown that these parameters can leak sensitive information in certain scenarios, such as with image data \citep{Le2018PPD}. To address this, exploring differential privacy techniques \citep{Dwork2014DP} to enhance the protection of shared model parameters is a critical avenue for future work. 

\paragraph{Personalized Learning.}
Heterogeneous settings are more applicable to many real-world problems, as clients often differ in computational capabilities, communication bandwidth, and local data distributions. These variations pose significant challenges for model convergence and performance. Thus, we propose the first methodology for learning Personalized DBNs by using the proximal operator. This framework can be easily extended to non-temporal cases, such as NOTEARS \citep{Zheng2018}. As mentioned earlier, many potential personalized federated learning frameworks can be applied, such as meta-learning and clustered federated learning. Moreover, there are several interesting research directions to explore. In our paper, we study the personalized setting under a fixed \( d \), i.e., the same number of variables for each client. A promising future direction is to address scenarios where \( d \) varies across different clients. Similarly, for \( p \), each client may have different time-lagged dependencies they wish to explore. Additionally, the case of mixed connectivity levels is an important research direction. For instance, in the presence of sparse networks, results for high-connectivity networks may be confounded, leading to missing edges. Several personalized federated learning methods have been proposed to address such issues. For example, FedNova \citep{Wang2020FedNova} normalizes aggregated updates to mitigate the impact of heterogeneous local computational workloads. More importantly, asynchronous techniques are particularly important for heterogeneous cases. Even though our asynchronous technique does not lead to a loss of accuracy, applying asynchronous methods remains crucial and represents a promising research direction, especially in the context of reinforcement learning, as explored by \citet{Lasse2018}.

\paragraph{Nonlinear Dependencies.}
Finally, we emphasize that the linear assumption of our methodology was made purely for simplicity, allowing us to focus on the most salient dynamic and temporal aspects of the problem. It is possible to model more complex nonlinear dependencies using Gaussian Processes \citep{jin2019physics, shen2023multi, gnanasambandam2024deep, yue2024federated, liu2024statistical} or neural networks. Additionally, the least squares loss function can be replaced with logistic loss (or more generally, any exponential family log-likelihood) to model binary data. Further, it would be valuable to consider combinations of continuous and discrete data \citep{Andrews2019LHd}, which are essential for many real-world applications.

\bibliographystyle{ACM-Reference-Format}
\bibliography{Reference}

%%% -*-BibTeX-*-
%%% Do NOT edit. File created by BibTeX with style
%%% ACM-Reference-Format-Journals [18-Jan-2012].

\begin{thebibliography}{63}

%%% ====================================================================
%%% NOTE TO THE USER: you can override these defaults by providing
%%% customized versions of any of these macros before the \bibliography
%%% command.  Each of them MUST provide its own final punctuation,
%%% except for \shownote{}, \showDOI{}, and \showURL{}.  The latter two
%%% do not use final punctuation, in order to avoid confusing it with
%%% the Web address.
%%%
%%% To suppress output of a particular field, define its macro to expand
%%% to an empty string, or better, \unskip, like this:
%%%
%%% \newcommand{\showDOI}[1]{\unskip}   % LaTeX syntax
%%%
%%% \def \showDOI #1{\unskip}           % plain TeX syntax
%%%
%%% ====================================================================

\ifx \showCODEN    \undefined \def \showCODEN     #1{\unskip}     \fi
\ifx \showDOI      \undefined \def \showDOI       #1{#1}\fi
\ifx \showISBNx    \undefined \def \showISBNx     #1{\unskip}     \fi
\ifx \showISBNxiii \undefined \def \showISBNxiii  #1{\unskip}     \fi
\ifx \showISSN     \undefined \def \showISSN      #1{\unskip}     \fi
\ifx \showLCCN     \undefined \def \showLCCN      #1{\unskip}     \fi
\ifx \shownote     \undefined \def \shownote      #1{#1}          \fi
\ifx \showarticletitle \undefined \def \showarticletitle #1{#1}   \fi
\ifx \showURL      \undefined \def \showURL       {\relax}        \fi
% The following commands are used for tagged output and should be
% invisible to TeX
\providecommand\bibfield[2]{#2}
\providecommand\bibinfo[2]{#2}
\providecommand\natexlab[1]{#1}
\providecommand\showeprint[2][]{arXiv:#2}

\bibitem[hip(2003)]%
        {hipaa_privacy_rule}
 \bibinfo{year}{2003}\natexlab{}.
\newblock \bibinfo{title}{HIPAA Privacy Rule: Summary of the Privacy Rule}.
\newblock \bibinfo{howpublished}{Code of Federal Regulations, 45 CFR Parts 160 and 164}.
\newblock
\newblock
\shownote{Available at: \url{https://www.hhs.gov/hipaa/for-professionals/privacy/laws-regulations/index.html}}.


\bibitem[Andrews et~al\mbox{.}(2019)]%
        {Andrews2019LHd}
\bibfield{author}{\bibinfo{person}{Bryan Andrews}, \bibinfo{person}{Joseph Ramsey}, {and} \bibinfo{person}{Gregory~F. Cooper}.} \bibinfo{year}{2019}\natexlab{}.
\newblock \showarticletitle{Learning High-dimensional Directed Acyclic Graphs with Mixed Data-types}. In \bibinfo{booktitle}{\emph{Proceedings of Machine Learning Research}} \emph{(\bibinfo{series}{Proceedings of Machine Learning Research}, Vol.~\bibinfo{volume}{104})}. \bibinfo{publisher}{PMLR}, \bibinfo{pages}{4--21}.
\newblock
\urldef\tempurl%
\url{https://proceedings.mlr.press/v104/andrews19a.html}
\showURL{%
\tempurl}


\bibitem[Arabi and Fang(2024)]%
        {arabi2024federated}
\bibfield{author}{\bibinfo{person}{Madi Arabi} {and} \bibinfo{person}{Xiaolei Fang}.} \bibinfo{year}{2024}\natexlab{}.
\newblock \showarticletitle{A Federated Data Fusion-Based Prognostic Model for Applications with Multi-Stream Incomplete Signals}.
\newblock \bibinfo{journal}{\emph{IISE Transactions}} \bibinfo{number}{just-accepted} (\bibinfo{year}{2024}), \bibinfo{pages}{1--56}.
\newblock


\bibitem[Benjamini and Hochberg(1995)]%
        {benjamini1995fdr}
\bibfield{author}{\bibinfo{person}{Yoav Benjamini} {and} \bibinfo{person}{Yosef Hochberg}.} \bibinfo{year}{1995}\natexlab{}.
\newblock \showarticletitle{Controlling the false discovery rate: a practical and powerful approach to multiple testing}.
\newblock \bibinfo{journal}{\emph{Journal of the Royal Statistical Society: Series B (Methodological)}} \bibinfo{volume}{57}, \bibinfo{number}{1} (\bibinfo{year}{1995}), \bibinfo{pages}{289--300}.
\newblock


\bibitem[Boyd et~al\mbox{.}(2011)]%
        {Boyd2011ADMM}
\bibfield{author}{\bibinfo{person}{Stephen Boyd}, \bibinfo{person}{Neal Parikh}, \bibinfo{person}{Eric Chu}, \bibinfo{person}{Borja Peleato}, {and} \bibinfo{person}{Jonathan Eckstein}.} \bibinfo{year}{2011}\natexlab{}.
\newblock \showarticletitle{Distributed Optimization and Statistical Learning via the Alternating Direction Method of Multipliers}.
\newblock \bibinfo{journal}{\emph{Foundations and Trends® in Machine Learning}} \bibinfo{volume}{3}, \bibinfo{number}{1} (\bibinfo{year}{2011}), \bibinfo{pages}{1--122}.
\newblock


\bibitem[Byrd et~al\mbox{.}(2003)]%
        {Byrd2003}
\bibfield{author}{\bibinfo{person}{R.~H. Byrd}, \bibinfo{person}{P. Lu}, \bibinfo{person}{J. Nocedal}, {and} \bibinfo{person}{C. Zhu}.} \bibinfo{year}{2003}\natexlab{}.
\newblock \showarticletitle{A Limited Memory Algorithm for Bound Constrained Optimization}.
\newblock \bibinfo{journal}{\emph{SIAM Journal on Scientific Computing}}  \bibinfo{volume}{16} (\bibinfo{year}{2003}), \bibinfo{pages}{1190--1208}.
\newblock
\urldef\tempurl%
\url{https://doi.org/10.1137/0916069}
\showDOI{\tempurl}


\bibitem[Chai et~al\mbox{.}(2020)]%
        {Chai2020SecureFederated}
\bibfield{author}{\bibinfo{person}{D. Chai}, \bibinfo{person}{L. Wang}, \bibinfo{person}{K. Chen}, {and} \bibinfo{person}{Q. Yang}.} \bibinfo{year}{2020}\natexlab{}.
\newblock \showarticletitle{Secure Federated Matrix Factorization}.
\newblock \bibinfo{journal}{\emph{IEEE Intelligent Systems}} \bibinfo{volume}{35}, \bibinfo{number}{1} (\bibinfo{date}{August} \bibinfo{year}{2020}), \bibinfo{pages}{30--38}.
\newblock


\bibitem[Chen et~al\mbox{.}(2018)]%
        {Chen2018FederatedMW}
\bibfield{author}{\bibinfo{person}{Fei Chen}, \bibinfo{person}{Mi Luo}, \bibinfo{person}{Zhenhua Dong}, \bibinfo{person}{Zhenguo Li}, {and} \bibinfo{person}{Xiuqiang He}.} \bibinfo{year}{2018}\natexlab{}.
\newblock \showarticletitle{Federated Meta-Learning with Fast Convergence and Efficient Communication}.
\newblock \bibinfo{journal}{\emph{arXiv: Learning}} (\bibinfo{year}{2018}).
\newblock
\urldef\tempurl%
\url{https://api.semanticscholar.org/CorpusID:209376818}
\showURL{%
\tempurl}


\bibitem[Chockalingam et~al\mbox{.}(2017)]%
        {Chock2017}
\bibfield{author}{\bibinfo{person}{Sabarathinam Chockalingam}, \bibinfo{person}{Wolter Pieters}, \bibinfo{person}{André Teixeira}, {and} \bibinfo{person}{P.H.A.J.M. Gelder}.} \bibinfo{year}{2017}\natexlab{}.
\newblock \bibinfo{booktitle}{\emph{Bayesian Network Models in Cyber Security: A Systematic Review}}.
\newblock \bibinfo{pages}{105--122}.
\newblock
\showISBNx{978-3-319-70289-6}
\urldef\tempurl%
\url{https://doi.org/10.1007/978-3-319-70290-2_7}
\showDOI{\tempurl}


\bibitem[Combettes and Pesquet(2011)]%
        {Combettes2011}
\bibfield{author}{\bibinfo{person}{P.~L. Combettes} {and} \bibinfo{person}{J.-C. Pesquet}.} \bibinfo{year}{2011}\natexlab{}.
\newblock \showarticletitle{Proximal Splitting Methods in Signal Processing}.
\newblock In \bibinfo{booktitle}{\emph{Fixed-Point Algorithms for Inverse Problems in Science and Engineering}}, \bibfield{editor}{\bibinfo{person}{H.~H. Bauschke}, \bibinfo{person}{R.~S. Burachik}, \bibinfo{person}{P.~L. Combettes}, \bibinfo{person}{V.~Elser}, \bibinfo{person}{D.~R. Luke}, {and} \bibinfo{person}{H.~Wolkowicz}} (Eds.). \bibinfo{publisher}{Springer New York}, \bibinfo{pages}{185--212}.
\newblock
\urldef\tempurl%
\url{https://doi.org/10.1007/978-1-4419-9569-8_10}
\showDOI{\tempurl}


\bibitem[Dwork and Roth(2014)]%
        {Dwork2014DP}
\bibfield{author}{\bibinfo{person}{Cynthia Dwork} {and} \bibinfo{person}{Aaron Roth}.} \bibinfo{year}{2014}\natexlab{}.
\newblock \showarticletitle{The Algorithmic Foundations of Differential Privacy}.
\newblock \bibinfo{journal}{\emph{Found. Trends Theor. Comput. Sci.}} \bibinfo{volume}{9}, \bibinfo{number}{3–4} (\bibinfo{date}{Aug.} \bibinfo{year}{2014}), \bibinfo{pages}{211–407}.
\newblock
\showISSN{1551-305X}
\urldef\tempurl%
\url{https://doi.org/10.1561/0400000042}
\showDOI{\tempurl}


\bibitem[Erdős and Rényi(1960)]%
        {ErdosRenyi1960}
\bibfield{author}{\bibinfo{person}{Paul Erdős} {and} \bibinfo{person}{Alfréd Rényi}.} \bibinfo{year}{1960}\natexlab{}.
\newblock \showarticletitle{On the Evolution of Random Graphs}.
\newblock \bibinfo{journal}{\emph{Publications of the Mathematical Institute of the Hungarian Academy of Sciences}}  \bibinfo{volume}{5} (\bibinfo{year}{1960}), \bibinfo{pages}{17--61}.
\newblock


\bibitem[Espeholt et~al\mbox{.}(2018)]%
        {Lasse2018}
\bibfield{author}{\bibinfo{person}{Lasse Espeholt}, \bibinfo{person}{Hubert Soyer}, \bibinfo{person}{Remi Munos}, \bibinfo{person}{Karen Simonyan}, \bibinfo{person}{Volodymir Mnih}, \bibinfo{person}{Tom Ward}, \bibinfo{person}{Yotam Doron}, \bibinfo{person}{Vlad Firoiu}, \bibinfo{person}{Tim Harley}, \bibinfo{person}{Iain Dunning}, \bibinfo{person}{Shane Legg}, {and} \bibinfo{person}{Koray Kavukcuoglu}.} \bibinfo{year}{2018}\natexlab{}.
\newblock \showarticletitle{IMPALA: Scalable Distributed Deep-RL with Importance Weighted Actor-Learner Architectures}.
\newblock  (\bibinfo{date}{02} \bibinfo{year}{2018}).
\newblock
\urldef\tempurl%
\url{https://doi.org/10.48550/arXiv.1802.01561}
\showDOI{\tempurl}


\bibitem[Friedman et~al\mbox{.}(1998)]%
        {Firedman2013}
\bibfield{author}{\bibinfo{person}{Nir Friedman}, \bibinfo{person}{Kevin Murphy}, {and} \bibinfo{person}{Stuart Russell}.} \bibinfo{year}{1998}\natexlab{}.
\newblock \showarticletitle{Learning the structure of dynamic probabilistic networks}. In \bibinfo{booktitle}{\emph{Proceedings of the Fourteenth Conference on Uncertainty in Artificial Intelligence}} (Madison, Wisconsin) \emph{(\bibinfo{series}{UAI'98})}. \bibinfo{publisher}{Morgan Kaufmann Publishers Inc.}, \bibinfo{address}{San Francisco, CA, USA}, \bibinfo{pages}{139–147}.
\newblock
\showISBNx{155860555X}


\bibitem[Ghosh et~al\mbox{.}(2022)]%
        {GhoshCFL2022}
\bibfield{author}{\bibinfo{person}{Avishek Ghosh}, \bibinfo{person}{Jichan Chung}, \bibinfo{person}{Dong Yin}, {and} \bibinfo{person}{Kannan Ramchandran}.} \bibinfo{year}{2022}\natexlab{}.
\newblock \showarticletitle{An Efficient Framework for Clustered Federated Learning}.
\newblock \bibinfo{journal}{\emph{IEEE Transactions on Information Theory}} \bibinfo{volume}{68}, \bibinfo{number}{12} (\bibinfo{year}{2022}), \bibinfo{pages}{8076--8091}.
\newblock
\urldef\tempurl%
\url{https://doi.org/10.1109/TIT.2022.3192506}
\showDOI{\tempurl}


\bibitem[Glymour et~al\mbox{.}(2019)]%
        {glymour2019review}
\bibfield{author}{\bibinfo{person}{Clark Glymour}, \bibinfo{person}{Kun Zhang}, {and} \bibinfo{person}{Peter Spirtes}.} \bibinfo{year}{2019}\natexlab{}.
\newblock \showarticletitle{Review of causal discovery methods based on graphical models}.
\newblock \bibinfo{journal}{\emph{Frontiers in Genetics}}  \bibinfo{volume}{10} (\bibinfo{year}{2019}), \bibinfo{pages}{524}.
\newblock


\bibitem[Gnanasambandam et~al\mbox{.}(2024)]%
        {gnanasambandam2024deep}
\bibfield{author}{\bibinfo{person}{Raghav Gnanasambandam}, \bibinfo{person}{Bo Shen}, \bibinfo{person}{Andrew Chung~Chee Law}, \bibinfo{person}{Chaoran Dou}, {and} \bibinfo{person}{Zhenyu Kong}.} \bibinfo{year}{2024}\natexlab{}.
\newblock \showarticletitle{Deep Gaussian process for enhanced Bayesian optimization and its application in additive manufacturing}.
\newblock \bibinfo{journal}{\emph{IISE Transactions}} (\bibinfo{year}{2024}), \bibinfo{pages}{1--14}.
\newblock


\bibitem[Gong et~al\mbox{.}(2023)]%
        {gong2023causal}
\bibfield{author}{\bibinfo{person}{Chang Gong}, \bibinfo{person}{Di Yao}, \bibinfo{person}{Chuzhe Zhang}, \bibinfo{person}{Wenbin Li}, {and} \bibinfo{person}{Jingping Bi}.} \bibinfo{year}{2023}\natexlab{}.
\newblock \showarticletitle{Causal Discovery from Temporal Data: An Overview and New Perspectives}.
\newblock \bibinfo{journal}{\emph{Journal of Artificial Intelligence Research}}  \bibinfo{volume}{75} (\bibinfo{year}{2023}), \bibinfo{pages}{231--268}.
\newblock


\bibitem[Gou et~al\mbox{.}(2007)]%
        {Guo2007}
\bibfield{author}{\bibinfo{person}{Kui~Xiang Gou}, \bibinfo{person}{Gong~Xiu Jun}, {and} \bibinfo{person}{Zheng Zhao}.} \bibinfo{year}{2007}\natexlab{}.
\newblock \showarticletitle{Learning Bayesian Network Structure from Distributed Homogeneous Data}. In \bibinfo{booktitle}{\emph{Eighth ACIS International Conference on Software Engineering, Artificial Intelligence, Networking, and Parallel/Distributed Computing (SNPD 2007)}}, Vol.~\bibinfo{volume}{3}. \bibinfo{pages}{250--254}.
\newblock
\urldef\tempurl%
\url{https://doi.org/10.1109/SNPD.2007.472}
\showDOI{\tempurl}


\bibitem[Huang et~al\mbox{.}(2015)]%
        {Huang2015ITD}
\bibfield{author}{\bibinfo{person}{Biwei Huang}, \bibinfo{person}{Kun Zhang}, {and} \bibinfo{person}{Bernhard Sch\"{o}lkopf}.} \bibinfo{year}{2015}\natexlab{}.
\newblock \showarticletitle{Identification of Time-Dependent Causal Model: a gaussian process treatment}. In \bibinfo{booktitle}{\emph{Proceedings of the 24th International Conference on Artificial Intelligence}} (Buenos Aires, Argentina) \emph{(\bibinfo{series}{IJCAI'15})}. \bibinfo{publisher}{AAAI Press}, \bibinfo{pages}{3561–3568}.
\newblock
\showISBNx{9781577357384}


\bibitem[Jiang et~al\mbox{.}(2019)]%
        {Jiang2019metaFL}
\bibfield{author}{\bibinfo{person}{Yihan Jiang}, \bibinfo{person}{Jakub Konečný}, \bibinfo{person}{Keith Rush}, {and} \bibinfo{person}{Sreeram Kannan}.} \bibinfo{year}{2019}\natexlab{}.
\newblock \bibinfo{title}{Improving Federated Learning Personalization via Model Agnostic Meta Learning}.
\newblock
\newblock
\urldef\tempurl%
\url{https://doi.org/10.48550/arXiv.1909.12488}
\showDOI{\tempurl}


\bibitem[Jin et~al\mbox{.}(2019)]%
        {jin2019physics}
\bibfield{author}{\bibinfo{person}{Xiaoning Jin}, \bibinfo{person}{Jun Ni}, {et~al\mbox{.}}} \bibinfo{year}{2019}\natexlab{}.
\newblock \showarticletitle{Physics-based Gaussian process for the health monitoring for a rolling bearing}.
\newblock \bibinfo{journal}{\emph{Acta astronautica}}  \bibinfo{volume}{154} (\bibinfo{year}{2019}), \bibinfo{pages}{133--139}.
\newblock


\bibitem[Khanna and Tan(2019)]%
        {Khanna2019EconomySR}
\bibfield{author}{\bibinfo{person}{Saurabh Khanna} {and} \bibinfo{person}{Vincent Yan~Fu Tan}.} \bibinfo{year}{2019}\natexlab{}.
\newblock \showarticletitle{Economy Statistical Recurrent Units For Inferring Nonlinear Granger Causality}.
\newblock \bibinfo{journal}{\emph{ArXiv}}  \bibinfo{volume}{abs/1911.09879} (\bibinfo{year}{2019}).
\newblock
\urldef\tempurl%
\url{https://api.semanticscholar.org/CorpusID:208248131}
\showURL{%
\tempurl}


\bibitem[Kontar et~al\mbox{.}(2021)]%
        {Kontar2021Ioft}
\bibfield{author}{\bibinfo{person}{Raed Kontar}, \bibinfo{person}{Naichen Shi}, \bibinfo{person}{Xubo Yue}, \bibinfo{person}{Seokhyun Chung}, \bibinfo{person}{Eunshin Byon}, \bibinfo{person}{Mosharaf Chowdhury}, \bibinfo{person}{Jionghua Jin}, \bibinfo{person}{Wissam Kontar}, \bibinfo{person}{Neda Masoud}, \bibinfo{person}{Maher Nouiehed}, \bibinfo{person}{Chinedum~E. Okwudire}, \bibinfo{person}{Garvesh Raskutti}, \bibinfo{person}{Romesh Saigal}, \bibinfo{person}{Karandeep Singh}, {and} \bibinfo{person}{Zhi-Sheng Ye}.} \bibinfo{year}{2021}\natexlab{}.
\newblock \showarticletitle{The Internet of Federated Things (IoFT)}.
\newblock \bibinfo{journal}{\emph{IEEE Access}}  \bibinfo{volume}{9} (\bibinfo{year}{2021}), \bibinfo{pages}{156071--156113}.
\newblock
\urldef\tempurl%
\url{https://doi.org/10.1109/ACCESS.2021.3127448}
\showDOI{\tempurl}


\bibitem[Lemoine et~al\mbox{.}(2021)]%
        {lemoine2021gwena}
\bibfield{author}{\bibinfo{person}{Guillaume~G. Lemoine}, \bibinfo{person}{Marie-Pier Scott-Boyer}, \bibinfo{person}{Baptiste Ambroise}, {et~al\mbox{.}}} \bibinfo{year}{2021}\natexlab{}.
\newblock \showarticletitle{{GWENA: gene co-expression networks analysis and extended modules characterization in a single Bioconductor package}}.
\newblock \bibinfo{journal}{\emph{BMC Bioinformatics}}  \bibinfo{volume}{22} (\bibinfo{year}{2021}), \bibinfo{pages}{267}.
\newblock
\urldef\tempurl%
\url{https://doi.org/10.1186/s12859-021-04179-4}
\showDOI{\tempurl}


\bibitem[Li et~al\mbox{.}(2021)]%
        {tian2021ditto}
\bibfield{author}{\bibinfo{person}{Tian Li}, \bibinfo{person}{Shengyuan Hu}, \bibinfo{person}{Ahmad Beirami}, {and} \bibinfo{person}{Virginia Smith}.} \bibinfo{year}{2021}\natexlab{}.
\newblock \showarticletitle{Ditto: Fair and Robust Federated Learning Through Personalization}. In \bibinfo{booktitle}{\emph{Proceedings of the 38th International Conference on Machine Learning}} \emph{(\bibinfo{series}{Proceedings of Machine Learning Research}, Vol.~\bibinfo{volume}{139})}, \bibfield{editor}{\bibinfo{person}{Marina Meila} {and} \bibinfo{person}{Tong Zhang}} (Eds.). \bibinfo{publisher}{PMLR}, \bibinfo{pages}{6357--6368}.
\newblock
\urldef\tempurl%
\url{https://proceedings.mlr.press/v139/li21h.html}
\showURL{%
\tempurl}


\bibitem[Li et~al\mbox{.}(2020)]%
        {Li2020FederatedLearning}
\bibfield{author}{\bibinfo{person}{Tian Li}, \bibinfo{person}{Ananda~K. Sahu}, \bibinfo{person}{Ameet Talwalkar}, {and} \bibinfo{person}{Virginia Smith}.} \bibinfo{year}{2020}\natexlab{}.
\newblock \showarticletitle{Federated Learning: Challenges, Methods, and Future Directions}.
\newblock \bibinfo{journal}{\emph{IEEE Signal Processing Magazine}} \bibinfo{volume}{37}, \bibinfo{number}{3} (\bibinfo{date}{May} \bibinfo{year}{2020}), \bibinfo{pages}{50--60}.
\newblock
\urldef\tempurl%
\url{https://doi.org/10.1109/MSP.2020.2978741}
\showDOI{\tempurl}


\bibitem[Liu and Liu(2024)]%
        {liu2024statistical}
\bibfield{author}{\bibinfo{person}{Xiao Liu} {and} \bibinfo{person}{Xinchao Liu}.} \bibinfo{year}{2024}\natexlab{}.
\newblock \showarticletitle{A Statistical Machine Learning Approach for Adapting Reduced-Order Models using Projected Gaussian Process}.
\newblock \bibinfo{journal}{\emph{arXiv preprint arXiv:2410.14090}} (\bibinfo{year}{2024}).
\newblock


\bibitem[Lu et~al\mbox{.}(2021)]%
        {lu2021causal}
\bibfield{author}{\bibinfo{person}{Jing Lu}, \bibinfo{person}{Bianca Dumitrascu}, \bibinfo{person}{Ian~C. McDowell}, \bibinfo{person}{Byung-Jun Jo}, \bibinfo{person}{Andres Barrera}, \bibinfo{person}{Lichen~K. Hong}, {and} \bibinfo{person}{et al.}} \bibinfo{year}{2021}\natexlab{}.
\newblock \showarticletitle{Causal network inference from gene transcriptional time-series response to glucocorticoids}.
\newblock \bibinfo{journal}{\emph{PLoS Computational Biology}} \bibinfo{volume}{17}, \bibinfo{number}{1} (\bibinfo{year}{2021}), \bibinfo{pages}{e1008223}.
\newblock
\urldef\tempurl%
\url{https://doi.org/10.1371/journal.pcbi.1008223}
\showDOI{\tempurl}


\bibitem[Madakam et~al\mbox{.}(2015)]%
        {Madakam2015IoT}
\bibfield{author}{\bibinfo{person}{Somayya Madakam}, \bibinfo{person}{R Ramaswamy}, {and} \bibinfo{person}{Siddharth Tripathi}.} \bibinfo{year}{2015}\natexlab{}.
\newblock \showarticletitle{Internet of Things (IoT): A Literature Review}.
\newblock \bibinfo{journal}{\emph{Journal of Computer and Communications}}  \bibinfo{volume}{3} (\bibinfo{date}{04} \bibinfo{year}{2015}), \bibinfo{pages}{164--173}.
\newblock
\urldef\tempurl%
\url{https://doi.org/10.4236/jcc.2015.35021}
\showDOI{\tempurl}


\bibitem[Malinsky and Spirtes(2019)]%
        {malinsky19a}
\bibfield{author}{\bibinfo{person}{Daniel Malinsky} {and} \bibinfo{person}{Peter Spirtes}.} \bibinfo{year}{2019}\natexlab{}.
\newblock \showarticletitle{Learning the Structure of a Nonstationary Vector Autoregression}. In \bibinfo{booktitle}{\emph{Proceedings of the Twenty-Second International Conference on Artificial Intelligence and Statistics}} \emph{(\bibinfo{series}{Proceedings of Machine Learning Research}, Vol.~\bibinfo{volume}{89})}, \bibfield{editor}{\bibinfo{person}{Kamalika Chaudhuri} {and} \bibinfo{person}{Masashi Sugiyama}} (Eds.). \bibinfo{publisher}{PMLR}, \bibinfo{pages}{2986--2994}.
\newblock
\urldef\tempurl%
\url{https://proceedings.mlr.press/v89/malinsky19a.html}
\showURL{%
\tempurl}


\bibitem[Marbach et~al\mbox{.}(2009)]%
        {marbach2009generating}
\bibfield{author}{\bibinfo{person}{D. Marbach}, \bibinfo{person}{T. Schaffter}, \bibinfo{person}{C. Mattiussi}, {and} \bibinfo{person}{D. Floreano}.} \bibinfo{year}{2009}\natexlab{}.
\newblock \showarticletitle{Generating Realistic In Silico Gene Networks for Performance Assessment of Reverse Engineering Methods}.
\newblock \bibinfo{journal}{\emph{Journal of Computational Biology}} \bibinfo{volume}{16}, \bibinfo{number}{2} (\bibinfo{year}{2009}), \bibinfo{pages}{229--239}.
\newblock
\urldef\tempurl%
\url{https://infoscience.epfl.ch/record/128148}
\showURL{%
\tempurl}


\bibitem[Margolin et~al\mbox{.}(2006)]%
        {margolin2006aracne}
\bibfield{author}{\bibinfo{person}{Alexander~A. Margolin}, \bibinfo{person}{Ilya Nemenman}, \bibinfo{person}{Karen Basso}, \bibinfo{person}{Chris Wiggins}, \bibinfo{person}{Gustavo Stolovitzky}, \bibinfo{person}{Riccardo Dalla~Favera}, {and} \bibinfo{person}{Andrea Califano}.} \bibinfo{year}{2006}\natexlab{}.
\newblock \showarticletitle{ARACNE: an algorithm for the reconstruction of gene regulatory networks in a mammalian cellular context}.
\newblock \bibinfo{journal}{\emph{BMC Bioinformatics}}  \bibinfo{volume}{7} (\bibinfo{year}{2006}), \bibinfo{pages}{S7}.
\newblock


\bibitem[McMahan et~al\mbox{.}(2016)]%
        {McMahan2016CommunicationEfficientLO}
\bibfield{author}{\bibinfo{person}{H.~B. McMahan}, \bibinfo{person}{Eider Moore}, \bibinfo{person}{Daniel Ramage}, \bibinfo{person}{Seth Hampson}, {and} \bibinfo{person}{Blaise~Ag{\"u}era y Arcas}.} \bibinfo{year}{2016}\natexlab{}.
\newblock \showarticletitle{Communication-Efficient Learning of Deep Networks from Decentralized Data}. In \bibinfo{booktitle}{\emph{International Conference on Artificial Intelligence and Statistics}}.
\newblock
\urldef\tempurl%
\url{https://api.semanticscholar.org/CorpusID:14955348}
\showURL{%
\tempurl}


\bibitem[Muhammad et~al\mbox{.}(2020)]%
        {Muhammad2020FedFast}
\bibfield{author}{\bibinfo{person}{Khalil Muhammad}, \bibinfo{person}{Qinqin Wang}, \bibinfo{person}{Diarmuid O'Reilly-Morgan}, \bibinfo{person}{Elias Tragos}, \bibinfo{person}{Barry Smyth}, \bibinfo{person}{Neil Hurley}, \bibinfo{person}{James Geraci}, {and} \bibinfo{person}{Aonghus Lawlor}.} \bibinfo{year}{2020}\natexlab{}.
\newblock \showarticletitle{FedFast: Going Beyond Average for Faster Training of Federated Recommender Systems}. \bibinfo{pages}{1234--1242}.
\newblock
\urldef\tempurl%
\url{https://doi.org/10.1145/3394486.3403176}
\showDOI{\tempurl}


\bibitem[Murphy(2002)]%
        {Murphy2002}
\bibfield{author}{\bibinfo{person}{Kevin~P. Murphy}.} \bibinfo{year}{2002}\natexlab{}.
\newblock \emph{\bibinfo{title}{Dynamic Bayesian Networks: Representation, Inference and Learning}}.
\newblock Ph.D. Thesis. \bibinfo{school}{University of California, Berkeley}.
\newblock


\bibitem[Na and Yang(2010)]%
        {Na2010DistributedBN}
\bibfield{author}{\bibinfo{person}{Yong-chan Na} {and} \bibinfo{person}{Jihoon Yang}.} \bibinfo{year}{2010}\natexlab{}.
\newblock \showarticletitle{Distributed Bayesian network structure learning}. In \bibinfo{booktitle}{\emph{Proceedings of the 2010 IEEE International Symposium on Industrial Electronics (ISIE)}}. \bibinfo{publisher}{IEEE}, \bibinfo{pages}{1607--1611}.
\newblock
\urldef\tempurl%
\url{https://doi.org/10.1109/ISIE.2010.5636500}
\showDOI{\tempurl}


\bibitem[Nauta et~al\mbox{.}(2019)]%
        {make1010019}
\bibfield{author}{\bibinfo{person}{Meike Nauta}, \bibinfo{person}{Doina Bucur}, {and} \bibinfo{person}{Christin Seifert}.} \bibinfo{year}{2019}\natexlab{}.
\newblock \showarticletitle{Causal Discovery with Attention-Based Convolutional Neural Networks}.
\newblock \bibinfo{journal}{\emph{Machine Learning and Knowledge Extraction}} \bibinfo{volume}{1}, \bibinfo{number}{1} (\bibinfo{year}{2019}), \bibinfo{pages}{312--340}.
\newblock
\showISSN{2504-4990}
\urldef\tempurl%
\url{https://www.mdpi.com/2504-4990/1/1/19}
\showURL{%
\tempurl}


\bibitem[Ng and Zhang(2022)]%
        {ng2022federated}
\bibfield{author}{\bibinfo{person}{Ignavier Ng} {and} \bibinfo{person}{Kun Zhang}.} \bibinfo{year}{2022}\natexlab{}.
\newblock \showarticletitle{Towards Federated Bayesian Network Structure Learning with Continuous Optimization}. In \bibinfo{booktitle}{\emph{Proceedings of the 25th International Conference on Artificial Intelligence and Statistics (AISTATS)}}, Vol.~\bibinfo{volume}{151}. \bibinfo{publisher}{PMLR}, \bibinfo{address}{Valencia, Spain}.
\newblock


\bibitem[Pamfil et~al\mbox{.}(2020)]%
        {pamfil2020dynotears}
\bibfield{author}{\bibinfo{person}{Razvan Pamfil}, \bibinfo{person}{Stefan Bauer}, \bibinfo{person}{Bernhard Sch{\"o}lkopf}, {and} \bibinfo{person}{Joachim~M. Buhmann}.} \bibinfo{year}{2020}\natexlab{}.
\newblock \showarticletitle{{DYNOTEARS}: Structure Learning from Time-Series Data}. In \bibinfo{booktitle}{\emph{Proceedings of the 23rd International Conference on Artificial Intelligence and Statistics (AISTATS)}}. \bibinfo{publisher}{PMLR}.
\newblock
\urldef\tempurl%
\url{http://proceedings.mlr.press/v108/pamfil20a.html}
\showURL{%
\tempurl}


\bibitem[Parikh and Boyd(2014a)]%
        {Parikh2014}
\bibfield{author}{\bibinfo{person}{N. Parikh} {and} \bibinfo{person}{S. Boyd}.} \bibinfo{year}{2014}\natexlab{a}.
\newblock \showarticletitle{Proximal Algorithms}.
\newblock \bibinfo{journal}{\emph{Foundations and Trends in Optimization}} \bibinfo{volume}{1}, \bibinfo{number}{3} (\bibinfo{date}{January} \bibinfo{year}{2014}), \bibinfo{pages}{127--239}.
\newblock
\urldef\tempurl%
\url{https://doi.org/10.1561/2400000003}
\showDOI{\tempurl}


\bibitem[Parikh and Boyd(2014b)]%
        {Neal2014Prox}
\bibfield{author}{\bibinfo{person}{Neal Parikh} {and} \bibinfo{person}{Stephen Boyd}.} \bibinfo{year}{2014}\natexlab{b}.
\newblock \showarticletitle{Proximal Algorithms}.
\newblock \bibinfo{journal}{\emph{Found. Trends Optim.}} \bibinfo{volume}{1}, \bibinfo{number}{3} (\bibinfo{date}{Jan.} \bibinfo{year}{2014}), \bibinfo{pages}{127–239}.
\newblock
\showISSN{2167-3888}
\urldef\tempurl%
\url{https://doi.org/10.1561/2400000003}
\showDOI{\tempurl}


\bibitem[Penfold et~al\mbox{.}(2015)]%
        {penfold2015csi}
\bibfield{author}{\bibinfo{person}{Christopher~A. Penfold}, \bibinfo{person}{Ahamed Shifaz}, \bibinfo{person}{Philip~E. Brown}, \bibinfo{person}{Alexander Nicholson}, {and} \bibinfo{person}{David~L. Wild}.} \bibinfo{year}{2015}\natexlab{}.
\newblock \showarticletitle{CSI: a nonparametric Bayesian approach to network inference from multiple perturbed time series gene expression data}.
\newblock \bibinfo{journal}{\emph{Statistical Applications in Genetics and Molecular Biology}} \bibinfo{volume}{14}, \bibinfo{number}{3} (\bibinfo{date}{Jun} \bibinfo{year}{2015}), \bibinfo{pages}{307--310}.
\newblock
\urldef\tempurl%
\url{https://doi.org/10.1515/sagmb-2014-0082}
\showDOI{\tempurl}


\bibitem[Penfold and Wild(2011)]%
        {penfold2011gene}
\bibfield{author}{\bibinfo{person}{Christopher~A. Penfold} {and} \bibinfo{person}{David~L. Wild}.} \bibinfo{year}{2011}\natexlab{}.
\newblock \showarticletitle{How to infer gene networks from expression profiles, revisited}.
\newblock \bibinfo{journal}{\emph{Interface Focus}}  \bibinfo{volume}{1} (\bibinfo{year}{2011}), \bibinfo{pages}{857--870}.
\newblock


\bibitem[Phong et~al\mbox{.}(2018)]%
        {Le2018PPD}
\bibfield{author}{\bibinfo{person}{Le~Trieu Phong}, \bibinfo{person}{Yoshinori Aono}, \bibinfo{person}{Takuya Hayashi}, \bibinfo{person}{Lihua Wang}, {and} \bibinfo{person}{Shiho Moriai}.} \bibinfo{year}{2018}\natexlab{}.
\newblock \showarticletitle{Privacy-Preserving Deep Learning via Additively Homomorphic Encryption}.
\newblock \bibinfo{journal}{\emph{Trans. Info. For. Sec.}} \bibinfo{volume}{13}, \bibinfo{number}{5} (\bibinfo{date}{May} \bibinfo{year}{2018}), \bibinfo{pages}{1333–1345}.
\newblock
\showISSN{1556-6013}


\bibitem[Rau et~al\mbox{.}(2010)]%
        {rau2010empirical}
\bibfield{author}{\bibinfo{person}{Andrea Rau}, \bibinfo{person}{Florence Jaffrézic}, \bibinfo{person}{Jean-Louis Foulley}, {and} \bibinfo{person}{Rebecca~W. Doerge}.} \bibinfo{year}{2010}\natexlab{}.
\newblock \showarticletitle{An empirical Bayesian method for estimating biological networks from temporal microarray data}.
\newblock \bibinfo{journal}{\emph{Statistical Applications in Genetics and Molecular Biology}}  \bibinfo{volume}{9} (\bibinfo{year}{2010}), \bibinfo{pages}{Article 9}.
\newblock
\urldef\tempurl%
\url{https://doi.org/10.2202/1544-6115.1513}
\showDOI{\tempurl}
\newblock
\shownote{Epub 2010 Jan 15}.


\bibitem[Sahu et~al\mbox{.}(2018)]%
        {Sahu2018FederatedOI}
\bibfield{author}{\bibinfo{person}{Anit~Kumar Sahu}, \bibinfo{person}{Tian Li}, \bibinfo{person}{Maziar Sanjabi}, \bibinfo{person}{Manzil Zaheer}, \bibinfo{person}{Ameet Talwalkar}, {and} \bibinfo{person}{Virginia Smith}.} \bibinfo{year}{2018}\natexlab{}.
\newblock \showarticletitle{Federated Optimization in Heterogeneous Networks}.
\newblock \bibinfo{journal}{\emph{arXiv: Learning}} (\bibinfo{year}{2018}).
\newblock
\urldef\tempurl%
\url{https://api.semanticscholar.org/CorpusID:59316566}
\showURL{%
\tempurl}


\bibitem[Sattler et~al\mbox{.}(2020)]%
        {Sattler2020CFL}
\bibfield{author}{\bibinfo{person}{Felix Sattler}, \bibinfo{person}{Klaus-Robert Müller}, {and} \bibinfo{person}{Wojciech Samek}.} \bibinfo{year}{2020}\natexlab{}.
\newblock \showarticletitle{Clustered Federated Learning: Model-Agnostic Distributed Multitask Optimization Under Privacy Constraints}.
\newblock \bibinfo{journal}{\emph{IEEE Transactions on Neural Networks and Learning Systems}}  \bibinfo{volume}{PP} (\bibinfo{date}{08} \bibinfo{year}{2020}), \bibinfo{pages}{1--13}.
\newblock
\urldef\tempurl%
\url{https://doi.org/10.1109/TNNLS.2020.3015958}
\showDOI{\tempurl}


\bibitem[Shen et~al\mbox{.}(2023)]%
        {shen2023multi}
\bibfield{author}{\bibinfo{person}{Bo Shen}, \bibinfo{person}{Raghav Gnanasambandam}, \bibinfo{person}{Rongxuan Wang}, {and} \bibinfo{person}{Zhenyu~James Kong}.} \bibinfo{year}{2023}\natexlab{}.
\newblock \showarticletitle{Multi-task Gaussian process upper confidence bound for hyperparameter tuning and its application for simulation studies of additive manufacturing}.
\newblock \bibinfo{journal}{\emph{IISE Transactions}} \bibinfo{volume}{55}, \bibinfo{number}{5} (\bibinfo{year}{2023}), \bibinfo{pages}{496--508}.
\newblock


\bibitem[Smith et~al\mbox{.}(2011)]%
        {smith2011network}
\bibfield{author}{\bibinfo{person}{Stephen~M Smith}, \bibinfo{person}{Karla~L Miller}, \bibinfo{person}{Gholamreza Salimi-Khorshidi}, \bibinfo{person}{Mathew Webster}, \bibinfo{person}{Christian~F Beckmann}, \bibinfo{person}{Thomas~E Nichols}, \bibinfo{person}{Joseph~D Ramsey}, {and} \bibinfo{person}{Mark~W Woolrich}.} \bibinfo{year}{2011}\natexlab{}.
\newblock \showarticletitle{Network modelling methods for FMRI}.
\newblock \bibinfo{journal}{\emph{NeuroImage}} \bibinfo{volume}{54}, \bibinfo{number}{2} (\bibinfo{year}{2011}), \bibinfo{pages}{875--891}.
\newblock
\urldef\tempurl%
\url{https://doi.org/10.1016/j.neuroimage.2010.08.063}
\showDOI{\tempurl}


\bibitem[Song et~al\mbox{.}(2009)]%
        {LeTV2009}
\bibfield{author}{\bibinfo{person}{Le Song}, \bibinfo{person}{Mladen Kolar}, {and} \bibinfo{person}{Eric~P. Xing}.} \bibinfo{year}{2009}\natexlab{}.
\newblock \showarticletitle{Time-varying dynamic Bayesian networks}. In \bibinfo{booktitle}{\emph{Proceedings of the 22nd International Conference on Neural Information Processing Systems}} (Vancouver, British Columbia, Canada) \emph{(\bibinfo{series}{NIPS'09})}. \bibinfo{publisher}{Curran Associates Inc.}, \bibinfo{address}{Red Hook, NY, USA}, \bibinfo{pages}{1732–1740}.
\newblock
\showISBNx{9781615679119}


\bibitem[Stolovitzky et~al\mbox{.}(2007)]%
        {stolovitzky2007dialogue}
\bibfield{author}{\bibinfo{person}{Gustavo Stolovitzky}, \bibinfo{person}{Don Monroe}, {and} \bibinfo{person}{Andrea Califano}.} \bibinfo{year}{2007}\natexlab{}.
\newblock \showarticletitle{Dialogue on Reverse-Engineering Assessment and Methods: The DREAM of High-Throughput Pathway Inference}.
\newblock \bibinfo{journal}{\emph{Annals of the New York Academy of Sciences}}  \bibinfo{volume}{1115} (\bibinfo{year}{2007}), \bibinfo{pages}{11--22}.
\newblock


\bibitem[Stolovitzky et~al\mbox{.}(2009)]%
        {stolovitzky2009lessons}
\bibfield{author}{\bibinfo{person}{Gustavo Stolovitzky}, \bibinfo{person}{Robert~J Prill}, {and} \bibinfo{person}{Andrea Califano}.} \bibinfo{year}{2009}\natexlab{}.
\newblock \showarticletitle{Lessons from the DREAM2 Challenges}.
\newblock \bibinfo{journal}{\emph{Annals of the New York Academy of Sciences}}  \bibinfo{volume}{1158} (\bibinfo{year}{2009}), \bibinfo{pages}{159--195}.
\newblock


\bibitem[Sun et~al\mbox{.}(2020)]%
        {SUN2020577}
\bibfield{author}{\bibinfo{person}{Yanning Sun}, \bibinfo{person}{Wei Qin}, {and} \bibinfo{person}{Zilong Zhuang}.} \bibinfo{year}{2020}\natexlab{}.
\newblock \showarticletitle{Quality consistency analysis for complex assembly process based on Bayesian networks}.
\newblock \bibinfo{journal}{\emph{Procedia Manufacturing}}  \bibinfo{volume}{51} (\bibinfo{year}{2020}), \bibinfo{pages}{577--583}.
\newblock
\showISSN{2351-9789}
\urldef\tempurl%
\url{https://doi.org/10.1016/j.promfg.2020.10.081}
\showDOI{\tempurl}
\newblock
\shownote{30th International Conference on Flexible Automation and Intelligent Manufacturing (FAIM2021)}.


\bibitem[Tank et~al\mbox{.}(2022)]%
        {Tank2022}
\bibfield{author}{\bibinfo{person}{A. Tank}, \bibinfo{person}{I. Covert}, \bibinfo{person}{N. Foti}, \bibinfo{person}{A. Shojaie}, {and} \bibinfo{person}{E.~B. Fox}.} \bibinfo{year}{2022}\natexlab{}.
\newblock \showarticletitle{Neural Granger Causality}.
\newblock \bibinfo{journal}{\emph{IEEE Transactions on Pattern Analysis and Machine Intelligence}} \bibinfo{volume}{44}, \bibinfo{number}{8} (\bibinfo{year}{2022}), \bibinfo{pages}{4267--4279}.
\newblock
\urldef\tempurl%
\url{https://doi.org/10.1109/TPAMI.2021.3065601}
\showDOI{\tempurl}


\bibitem[Tsamardinos et~al\mbox{.}(2006)]%
        {tsamardinos2006max}
\bibfield{author}{\bibinfo{person}{Ioannis Tsamardinos}, \bibinfo{person}{Laura~E Brown}, {and} \bibinfo{person}{Constantin~F Aliferis}.} \bibinfo{year}{2006}\natexlab{}.
\newblock \showarticletitle{The max-min hill-climbing Bayesian network structure learning algorithm}.
\newblock \bibinfo{journal}{\emph{Machine learning}} \bibinfo{volume}{65}, \bibinfo{number}{1} (\bibinfo{year}{2006}), \bibinfo{pages}{31--78}.
\newblock


\bibitem[Wang et~al\mbox{.}(2020)]%
        {Wang2020FedNova}
\bibfield{author}{\bibinfo{person}{Jianyu Wang}, \bibinfo{person}{Zachary Charles}, \bibinfo{person}{Brendan Childers}, \bibinfo{person}{Yu-Xiang Sun}, \bibinfo{person}{Suhas Sreehari}, \bibinfo{person}{Sebastian~U Stich}, \bibinfo{person}{Mikhail Dmitriev}, \bibinfo{person}{Ameet Talwalkar}, {and} \bibinfo{person}{Michael~I Jordan}.} \bibinfo{year}{2020}\natexlab{}.
\newblock \showarticletitle{Federated learning with normalized averaging}. In \bibinfo{booktitle}{\emph{Neural Information Processing Systems (NeurIPS)}}.
\newblock


\bibitem[Yang et~al\mbox{.}(2019)]%
        {Yang2019FederatedMachine}
\bibfield{author}{\bibinfo{person}{Qiang Yang}, \bibinfo{person}{Yang Liu}, \bibinfo{person}{Tian Chen}, {and} \bibinfo{person}{Yongxin Tong}.} \bibinfo{year}{2019}\natexlab{}.
\newblock \showarticletitle{Federated Machine Learning: Concept and Applications}.
\newblock \bibinfo{journal}{\emph{ACM Transactions on Intelligent Systems and Technology (TIST)}} \bibinfo{volume}{10}, \bibinfo{number}{2} (\bibinfo{date}{January} \bibinfo{year}{2019}), \bibinfo{pages}{1--19}.
\newblock
\urldef\tempurl%
\url{https://doi.org/10.1145/3308558}
\showDOI{\tempurl}


\bibitem[Yu et~al\mbox{.}(2004)]%
        {Yu2004}
\bibfield{author}{\bibinfo{person}{Jing Yu}, \bibinfo{person}{V.~Anne Smith}, \bibinfo{person}{Paul~P. Wang}, \bibinfo{person}{Alexander~J. Hartemink}, {and} \bibinfo{person}{Erich~D. Jarvis}.} \bibinfo{year}{2004}\natexlab{}.
\newblock \showarticletitle{Advances to Bayesian Network Inference for Generating Causal Networks from Observational Biological Data}.
\newblock \bibinfo{journal}{\emph{Bioinformatics}} \bibinfo{volume}{20}, \bibinfo{number}{18} (\bibinfo{year}{2004}), \bibinfo{pages}{3594--3603}.
\newblock
\urldef\tempurl%
\url{https://doi.org/10.1093/bioinformatics/bth448}
\showDOI{\tempurl}


\bibitem[Yue and Kontar(2024)]%
        {yue2024federated}
\bibfield{author}{\bibinfo{person}{Xubo Yue} {and} \bibinfo{person}{Raed Kontar}.} \bibinfo{year}{2024}\natexlab{}.
\newblock \showarticletitle{Federated Gaussian Process: Convergence, Automatic Personalization and Multi-fidelity Modeling}.
\newblock \bibinfo{journal}{\emph{IEEE Transactions on Pattern Analysis and Machine Intelligence}} (\bibinfo{year}{2024}).
\newblock


\bibitem[Zeng et~al\mbox{.}(2016)]%
        {zeng2016discovering}
\bibfield{author}{\bibinfo{person}{Zhiya Zeng}, \bibinfo{person}{Xia Jiang}, {and} \bibinfo{person}{Richard Neapolitan}.} \bibinfo{year}{2016}\natexlab{}.
\newblock \showarticletitle{Discovering causal interactions using Bayesian network scoring and information gain}.
\newblock \bibinfo{journal}{\emph{BMC Bioinformatics}}  \bibinfo{volume}{17} (\bibinfo{year}{2016}), \bibinfo{pages}{221}.
\newblock
\urldef\tempurl%
\url{https://doi.org/10.1186/s12859-016-1084-8}
\showDOI{\tempurl}


\bibitem[Zhang et~al\mbox{.}(2024)]%
        {zhang2024federated}
\bibfield{author}{\bibinfo{person}{Zihan Zhang}, \bibinfo{person}{Shancong Mou}, \bibinfo{person}{Mostafa Reisi~Gahrooei}, \bibinfo{person}{Massimo Pacella}, {and} \bibinfo{person}{Jianjun Shi}.} \bibinfo{year}{2024}\natexlab{}.
\newblock \showarticletitle{Federated Multiple Tensor-on-Tensor Regression (FedMTOT) for Multimodal Data under Data-Sharing Constraints}.
\newblock \bibinfo{journal}{\emph{Technometrics}} (\bibinfo{year}{2024}), \bibinfo{pages}{1--26}.
\newblock


\bibitem[Zheng et~al\mbox{.}(2018)]%
        {Zheng2018}
\bibfield{author}{\bibinfo{person}{Xun Zheng}, \bibinfo{person}{Bryon Aragam}, \bibinfo{person}{Pradeep~K. Ravikumar}, {and} \bibinfo{person}{Eric~P. Xing}.} \bibinfo{year}{2018}\natexlab{}.
\newblock \showarticletitle{DAGs with NO TEARS: Continuous optimization for structure learning}. In \bibinfo{booktitle}{\emph{Advances in Neural Information Processing Systems 31}}. \bibinfo{publisher}{Curran Associates, Inc.}, \bibinfo{pages}{9472--9483}.
\newblock


\end{thebibliography}

\section{Supplementary Materials}
\subsection{Notation of multivariate time-series }
\label{m_not}
After reintroducing the index \( m \) for the \( M \) realizations, we can stack the data for client \( k \). For each client \( k \), define \( n_k = M (T + 1 - p) \) as the effective sample size. Then, we can write:

\[
X_t^k = X_t^k W + X_{t-1}^k A_1 + \dots + X_{t-p}^k A_p + Z^k,
\]
where:
\begin{itemize}
    \item \( X_t^k \) is a \( n_k \times d \) matrix whose rows are \((x_{m,t}^k)^\top \) for \( m = 1, \dots, M \);
    \item  \( X_{t-i}^k\) are similarly defined time-lagged matrices for \( i = 1, \dots, p \);
    \item \( Z^k \) aggregates the noise terms \((u_{m,t}^k)^\top\).
\end{itemize}

This formulation allows us to handle multiple realizations per client while maintaining the SVAR structure across time and variables. The remaining learning and optimization steps will follow exactly the approach outlined in the main paper.

\subsection{Simulation Data Generating:}
\label{sim}
\textbf{Intra-slice graph:} We use the \textit{Erdős-Rényi (ER) model} to generate a random, directed acyclic graph (DAG) with a target mean degree \(pr\). In the ER model, edges are generated independently using i.i.d. Bernoulli trials with a probability \(pr/dr\), where \(dr\) is the number of nodes. The resulting graph is first represented as an adjacency matrix and then oriented to ensure acyclicity by imposing a lower triangular structure, producing a valid DAG. Finally, the nodes of the DAG are randomly permuted to remove any trivial ordering, resulting in a randomized and realistic structure suitable for downstream applications.\\
\textbf{Inter-slice graph:} We still use \textit{ER model} to generate the weighted matrix. The edges are directed from node \( i_{t-1} \) at time \( t-1 \) to node \( j_t \) at time \( t \). The binary adjacency matrix \( A_{\text{bin}} \) is constructed as:
\[
A_{i_{t-1}, j_{t}} = 
\begin{cases}
1 & \text{with probability } pr/dr \quad \text{for edges from node } i_{t-1} \text{ to } j_t, \\
0 & \text{otherwise}.
\end{cases}
\]
\textbf{Assigning Weights}: 
Once the binary adjacency matrix is generated, we assign edge weights from a \textit{uniform distribution} over the range \([-0.5, -0.3] \cup [0.3, 0.5]\) for \(W\) and \([-0.5\alpha, -0.3\alpha] \cup [0.3\alpha, 0.5\alpha]\) for \(A\), where:
\[
\alpha = \frac{1}{\eta^{p-1}},
\]
and \( \eta \geq 1 \) is a decay parameter controlling how the influence of edges decreases as time steps get further apart.

\subsection{Hyperparameters analysis} 
\label{Hyper}
In this section, we present the optimal parameter values for each simulation. The following table records the optimal \(\lambda_a\) and \(\lambda_w\) values for experiments with varying numbers of variables (\(d\)). In general, \(\lambda_a = 0.5\) and \(\lambda_w = 0.5\) perform well in all cases. However, when \(\lambda_a, \lambda_w > 0.5\), the algorithm sometimes outputs zero matrices.  

\begin{table}[h!]
\centering
\caption{Optimal \(\lambda_a\) and \(\lambda_w\) values for varying \(d\).}
\begin{tabular}{lcccc}
\toprule
 & \textbf{\(d = 5\)} & \textbf{\(d = 10\)} & \textbf{\(d = 15\)} & \textbf{\(d = 20\)} \\ 
\midrule
\(\lambda_a\) & 0.5 & 0.5 & 0.5 & 0.5 \\ 
\(\lambda_w\) & 0.5 & 0.5 & 0.5 & 0.5 \\  
\bottomrule
\end{tabular}
\end{table}

The following tables summarize the optimal \(\lambda_a\) and \(\lambda_w\) values for experiments with varying numbers of clients, keeping \(n = 256\). In general, \(\lambda_a\) values between 0.4 and 0.5 and \(\lambda_w\) values between 0 and 0.5 work well. Notably, simulations indicate that the optimal range of \(\lambda_a\) and \(\lambda_w\) should be between 0.05 and 0.5, depending on the network topology. Within this range, the error in Structural Hamming Distance (SHD) is typically less than 5 units from the optimal value.

\begin{table}[h!]
\centering
\caption{\textbf{Sample Size with 256 and \(d = 10\)}}
\begin{tabular}{lcccccc}
\toprule
 & \textbf{\(k=2\)} & \textbf{\(k=4\)} & \textbf{\(k=8\)} & \textbf{\(k=16\)} & \textbf{\(k=32\)} & \textbf{\(k=64\)} \\ 
\midrule
\(\lambda_a\) & 0.05 & 0.5 & 0.5 & 0.4 & 0.35 & 0.4 \\ 
\(\lambda_w\) & 0.3 & 0.5 & 0.45 & 0.5 & 0.25 & 0.3  \\  
\bottomrule
\end{tabular}
\end{table}

\begin{table}[h!]
\centering
\caption{\textbf{Sample Size with 256 and \(d = 20\)}}
\begin{tabular}{lcccccc}
\toprule
 & \textbf{\(k=2\)} & \textbf{\(k=4\)} & \textbf{\(k=8\)} & \textbf{\(k=16\)} & \textbf{\(k=32\)} & \textbf{\(k=64\)} \\ 
\midrule
\(\lambda_a\) & 0.3 & 0.2 & 0.5 & 0.5 & 0.5 & 0.05 \\ 
\(\lambda_w\) & 0.4 & 0.5 & 0.5 & 0.5 & 0.5 & 0.25 \\ 
\bottomrule
\end{tabular}
\end{table}

\subsection{Closed form for $B_k$ and $D_k$}
\label{closed form}
Minimize with respect to \( B_k \) and \( D_k \):

\begin{align*}
J(B_k, D_k) = &\, \ell_k(B_k, D_k) + \operatorname{Tr}\left( \beta_k^t (B_k - W^{(t)})^\top \right) + \frac{\rho_2^t}{2} \left\| B_k - W^{(t)} \right\|_F^2 \\
& + \operatorname{Tr}\left( \gamma_k^t (D_k - A^{(t)})^\top \right) + \frac{\rho_2^t}{2} \left\| D_k - A^{(t)} \right\|_F^2
\end{align*}
where:
\[
\ell_k(B_k, D_k) = \frac{1}{2n} \left\| X_t - X_t B_k - X_{(t-p):(t-1)} D_k \right\|_F^2.
\]

Due to gradients and optimality conditions, we can set the gradients of \( J(B_k, D_k) \) with respect to \( B_k \) and \( D_k \) to zero. Thus we can have: 

\textbf{Gradient with respect to \( B_k \):}

\[
\nabla_{B_k} J = -\frac{1}{n} X_t^\top \left( X_t - X_t B_k - X_{(t-p):(t-1)} D_k \right) + \beta_k^t + \rho_2^t (B_k - W^{(t)}) = 0.
\]
Simplify:
\begin{align*}
&(-S + S B_k + M D_k) + \beta_k^t + \rho_2^t (B_k - W^{(t)}) = 0 \\
& \implies (S + \rho_2^t I) B_k + M D_k = S - \beta_k^t + \rho_2^t W^{(t)} \\
& \implies P B_k + M D_k = b_1.
\end{align*}
\textbf{Gradient with respect to \( D_k \):}
\[
\nabla_{D_k} J = -\frac{1}{n} X_{(t-p):(t-1)}^\top \left( X_t - X_t B_k - X_{(t-p):(t-1)} D_k \right) + \gamma_k^t + \rho_2^t (D_k - A^{(t)}) = 0.
\]
Simplify:
\begin{align*}
&(-M^\top + M^\top B_k + N D_k) + \gamma_k^t + \rho_2^t (D_k - A^{(t)}) = 0 \\
& \implies M^\top B_k + (N + \rho_2^t I) D_k = M^\top - \gamma_k^t + \rho_2^t A^{(t)}\\
& \implies M^\top B_k + Q D_k = b_2.
\end{align*}

\subsection{Closed form for personalized DBN learning}
\label{pclosed}
Update for \(\mathbf{B}_k\)

\begin{equation}
f_{\tilde{W}_k}(\tilde{W}_k)
\;=\;
\mu\|W_k^{(t+1)} - \tilde{W}_k\|^2
\;+\;
\mathrm{tr}\bigl(\beta_k^{(t)\top}(\tilde{W}_k - W^{(t)})\bigr)
\;+\;
\tfrac{\rho_2^{(t)}}{2}\,\|\tilde{W}_k - W^{(t)}\|^2
\end{equation}

Hence,
\[
\nabla_{\tilde{W}_k}\,f_{\tilde{W}_k}(\tilde{W}_k)
\;=\;
2\mu\,(\tilde{W}_k - W_k^{(t+1)}) 
\;+\; 
\beta_k^{(t)} 
\;+\;
\rho_2^{(t)}\,(\tilde{W}_k - W^{(t)}).
\]

By setting \(\nabla_{\tilde{W}_k} f_{\tilde{W}_k}(\tilde{W}_k) = 0\), we have 
\[
2\mu(\tilde{W}_k - W_k^{(t+1)}) \;+\; \beta_k^{(t)} 
\;+\; \rho_2^{(t)}(\tilde{W}_k - W^{(t)}) \;=\; 0.
\]

\[
\bigl(2\mu + \rho_2^{(t)}\bigr)\,\tilde{W}_k
\;=\;
2\mu\,W_k^{(t+1)}
\;+\;
\rho_2^{(t)}\,W^{(t)}
\;-\;
\beta_k^{(t)}.
\]

\[
\boxed{
\tilde{W}_k^{(t+1)} 
\;=\;
\frac{
  2\mu\,W_k^{(t+1)} 
  \;+\;
  \rho_2^{(t)}\,W^{(t)}
  \;-\;
  \beta_k^{(t)}
}{
  2\mu + \rho_2^{(t)}
}.
}
\]

Update for \(\mathbf{D}_k\)

\[
f_{\tilde{A}_k}(\tilde{A}_k)
\;=\;
\mu\|A_k^{(t+1)} - \tilde{A}_k\|^2
\;+\;
\mathrm{tr}\bigl(\gamma_k^{(t)\top}(\tilde{A}_k - A^{(t)})\bigr)
\;+\;
\tfrac{\rho_2^{(t)}}{2}\,\|\tilde{A}_k - A^{(t)}\|^2.
\]

Hence,
\[
\nabla_{\tilde{A}_k}\,f_{\tilde{A}_k}(\tilde{A}_k)
\;=\;
2\mu\,(\tilde{A}_k - A_k^{(t+1)})
\;+\;
\gamma_k^{(t)}
\;+\;
\rho_2^{(t)}\,(\tilde{A}_k - A^{(t)}).
\]

By \(\nabla_{\tilde{A}_k}\,f_{\tilde{A}_k}(\tilde{A}_k) = 0\), we have 
\[
2\mu(\tilde{A}_k - A_k^{(t+1)}) \;+\; \gamma_k^{(t)} 
\;+\; \rho_2^{(t)}(\tilde{A}_k - A^{(t)}) 
\;=\; 0.
\]

\[
\bigl(2\mu + \rho_2^{(t)}\bigr)\,\tilde{A}_k
\;=\;
2\mu\,A_k^{(t+1)}
\;+\;
\rho_2^{(t)}\,A^{(t)}
\;-\;
\gamma_k^{(t)}.
\]

\[
\boxed{
\tilde{A}_k^{(t+1)} 
\;=\;
\frac{
  2\mu\,A_k^{(t+1)} 
  \;+\;
  \rho_2^{(t)}\,A^{(t)}
  \;-\;
  \gamma_k^{(t)}
}{
  2\mu + \rho_2^{(t)}
}.
}
\]

\subsection{Application in FMRI}
\label{app_FMRI}
In this section, we present the recorded AUROC values for our \texttt{FDBNL} method, as shown in Table.\ref{fMRI_re}.  

\begin{table}[H]
\centering
\label{fMRI_re}
\begin{tabular}{lccccc}
\toprule
AUROC & 1 & 2 & 3 & 4 & 5 \\
\midrule
\( d = 15 \) & 0.68 & 0.75 & 0.71 & 0.77 & 0.77 \\
\( d = 10 \) & 0.69 & 0.77 & 0.71 & 0.69 & 0.83 \\
\( d = 5 \)  & 0.70 & 0.75 & 0.76 & 0.78 & 0.70 \\
\bottomrule
\end{tabular}
\end{table}

Our model outputs two matrices, \( W \) and \( A \), which represent strong connections and weak connections, respectively. To produce the final weight matrix, we combine these two matrices using an element-wise sum. We found that \(\lambda_W = 0.05\) and \(\lambda_A = 0.01\) work well across all datasets by several round experiment. 

The following table is for \texttt{PFDBNL}: 

\begin{table}[H]
\centering
\label{fMRI_re}
\begin{tabular}{lccccc}
\toprule
AUROC & 1 & 2 & 3 & 4 & 5   \\
\midrule
\( d = 15 \) & 0.67 & 0.77 & 0.71 & 0.78 & 0.77  \\
\( d = 10 \) & 0.78 & 0.71 & 0.70 & 0.76 & 0.75  \\
\( d = 5 \)  & 0.73 & 0.78 & 0.70 & 0.83 & 0.78   \\
\bottomrule
\end{tabular}
\end{table}

\subsection{Application in Dream4}

We have attached our AUPR and AUROC for each dataset at Tab.\ref{dream4_re}. Similar to the fMRI data, our model outputs two matrices, \( W \) and \( A \). However, these are interpreted as representing fast-acting and slow-acting influences, respectively. To produce the final weight matrix, we combine these two matrices using an element-wise sum. Based on the hyperparameter analysis by \citet{pamfil2020dynotears}, we found that \(\lambda_W = 0.0025\) and \(\lambda_A = 0.0025\) work well across all datasets.  

\begin{table}[h!]
\centering
\label{dream4_re}
\begin{tabular}{lcc}
\hline
\textbf{Dataset} & \textbf{AUPR} & \textbf{AUROC} \\ \hline
1                & 0.054         & 0.64          \\ \hline
2                & 0.032         & 0.58          \\ \hline
3                & 0.041         & 0.60          \\ \hline
4                & 0.034         & 0.58          \\ \hline
5                & 0.035         & 0.62          \\ \hline
\textbf{Mean ± Std} & 0.040 ± 0.008 & 0.60 ± 0.022 \\ \hline
\end{tabular}
\end{table}

\subsection{Result for \(A\)}
\label{result_a}

\begin{figure}[H]
\centering
\includegraphics[width=0.8\textwidth]{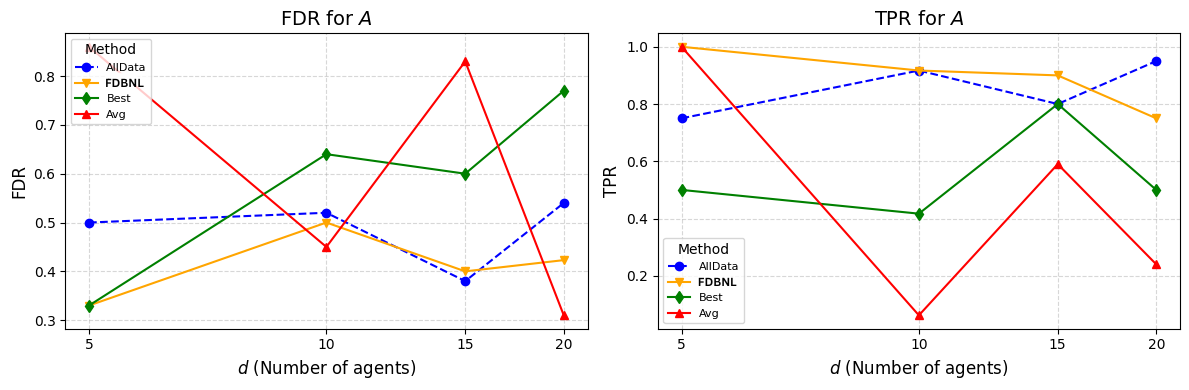}
\caption{Structure learning results for \(A\) in a DBN with Gaussian noise for \(d = 5, 10, 15, 20\) variables, an autoregressive order \(p = 1\), and \(K = 10\) clients. Each metric value indicates the mean performance across 10 different simulated datasets}
\label{vary_d_a}
\end{figure}

\begin{figure}[H]
\centering
\includegraphics[width=0.8\textwidth]{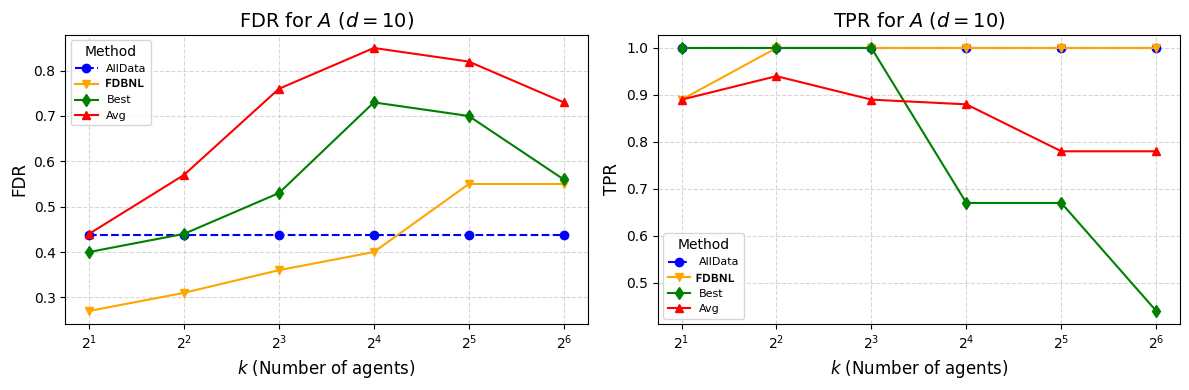}
\caption{Structure learning for $A$ of DBN with Gaussian noise for $d = 10$ variables, $p = 1$ Autoregressive order and varying number of clients. There are $n = 256$ samples in total, distributed evenly across $K \in \{2, 4, 8, 16, 32, 64\}$ clients. Each metric value indicates the mean performance across 10 different simulated datasets}
\label{10d_256client_a}
\end{figure}

\begin{figure}[H]
\centering
\includegraphics[width=0.8\textwidth]{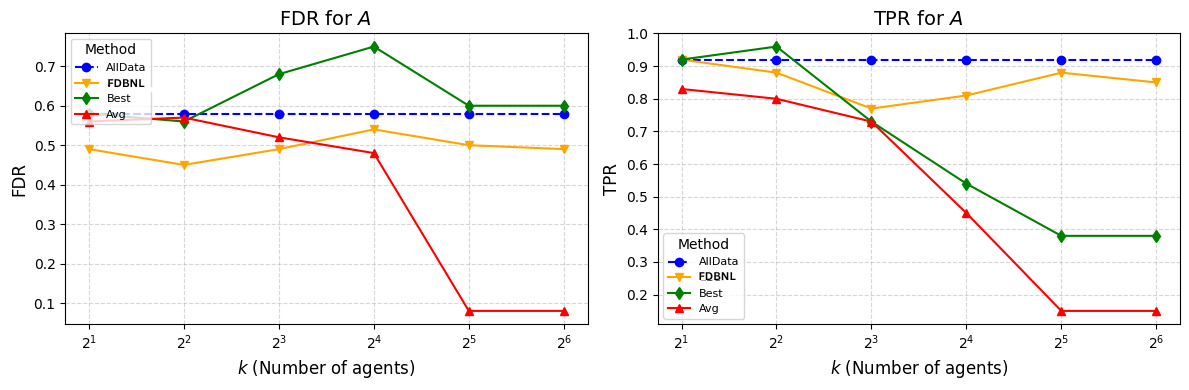}
\caption{Structure learning for $A$ of DBN with Gaussian noise for $d = 20$ variables, $p = 1$ Autoregressive order and varying number of clients. There are $n = 256$ samples in total, distributed evenly across $K \in \{2, 4, 8, 16, 32, 64\}$ clients. Each metric value indicates the mean performance across 10 different simulated datasets}
\label{20d_256client_a}
\end{figure}

\label{result_a}
\begin{figure}[H]
\centering
\includegraphics[width=0.8\textwidth]{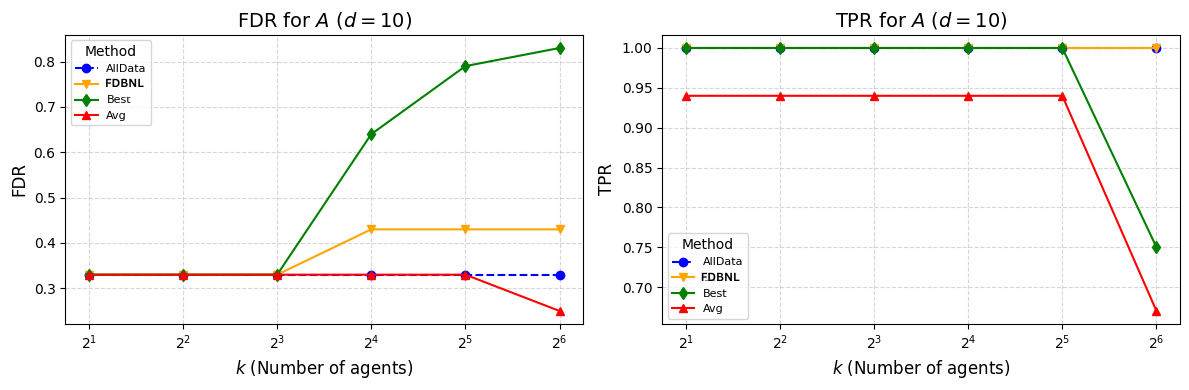}
\caption{Structure learning for $A$ of DBN with Gaussian noise for $d = 10$ variables, $p = 1$ Autoregressive order and varying number of clients. There are $n = 512$ samples in total, distributed evenly across $K \in \{2, 4, 8, 16, 32, 64\}$ clients. Each metric value indicates the mean performance across 10 different simulated datasets}
\label{10d_512client_a}
\end{figure}

\begin{figure}[H]
\centering
\includegraphics[width=0.8\textwidth]{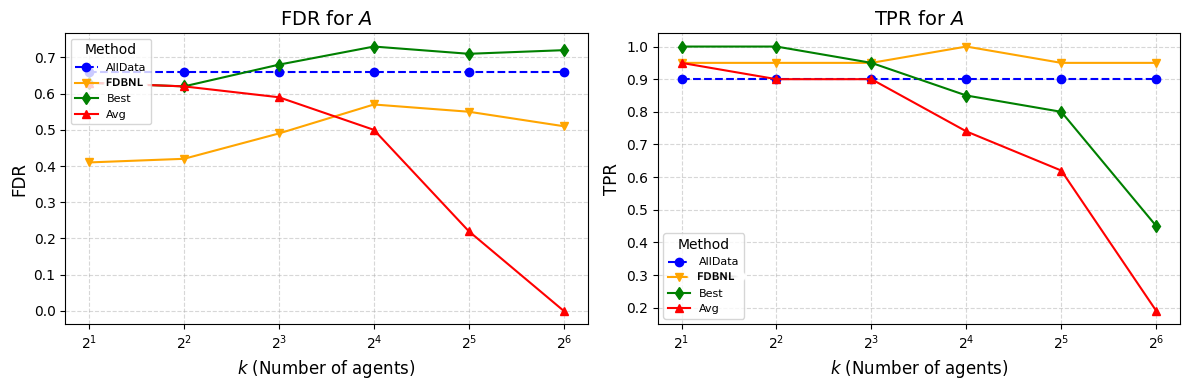}
\caption{Structure learning for $A$ of DBN with Gaussian noise for $d = 20$ variables, $p = 1$ Autoregressive order and varying number of clients. There are $n = 512$ samples in total, distributed evenly across $K \in \{2, 4, 8, 16, 32, 64\}$ clients. Each metric value indicates the mean performance across 10 different simulated datasets}
\label{20d_512client_a}
\end{figure}

\end{document}